\documentclass[journal]{IEEEtran}
\IEEEoverridecommandlockouts

\usepackage{amsfonts,amsmath,amssymb}
\usepackage{array}
    \newcolumntype{M}{>{\ttfamily}c}
\usepackage{booktabs,multirow}
\usepackage{bm}
\usepackage{cite}
\usepackage{graphicx}
\usepackage{textcomp}
\usepackage{algorithmic}
\usepackage{algorithm}
\usepackage{xcolor}
\usepackage{listings}

\definecolor{backcolour}{rgb}{0.95,0.95,0.92}
\definecolor{codegreen}{rgb}{0,0.6,0}
\definecolor{codegray}{rgb}{0.5,0.5,0.5}
\definecolor{codepurple}{rgb}{0.58,0,0.82}

\lstdefinestyle{mystyle}{
    backgroundcolor=\color{backcolour},
    commentstyle=\color{codegreen},
    keywordstyle=\color{magenta},
    numberstyle=\tiny\color{codegray},
    stringstyle=\color{codepurple},
    basicstyle=\ttfamily\footnotesize,
    breakatwhitespace=false,
    breaklines=true,
    captionpos=b,
    keepspaces=true,
    numbers=left,
    numbersep=5pt,
    showspaces=false,
    showstringspaces=false,
    showtabs=false,
    tabsize=2
}
\def\BibTeX{{\rm B\kern-.05em{\sc i\kern-.025em b}\kern-.08em
    T\kern-.1667em\lower.7ex\hbox{E}\kern-.125emX}}

\begin{document}

\author{
Jinsheng Yuan*,
Yuhang Hao,
Yun Wu,~\IEEEmembership{Member,~IEEE,}
Chongyan Gu,~\IEEEmembership{Senior Member,~IEEE,}
Weisi Guo,~\IEEEmembership{Senior Member,~IEEE}

\thanks{Jinsheng Yuan (\textsuperscript{*}Corresponding author) and Weisi Guo are with the Faculty of Engineering \& Applied Sciences, Cranfield University, UK, supported by EPSRC CHEDDAR: Communications Hub for Empowering Distributed clouD computing Applications and Research (EP/X040518/1, EP/Y037421/1). E-mail: \{Jinsheng.Yuan, Weisi.Guo\}@cranfield.ac.uk.}

\thanks{Yuhang Hao, Yun Wu and Chongyan Gu are with Queen's University Belfast, supported by EPSRC New Investigator Award (EP/X009602/1). E-mail: \{yhao09, yun.wu, C.Gu\}@qub.ac.uk)}
}

\title{Remote Rowhammer Attack using Adversarial Observations on Federated Learning Clients}

\maketitle

\begin{abstract}
Federated Learning (FL) has the potential for simultaneous global learning amongst a large number of parallel agents, enabling emerging AI systems to be trained across demographically diverse data. Central to this being efficient is the ability for FL to perform sparse gradient updates and remote direct memory access at the central server. Most of the research in FL security focuses on protecting data privacy at the edge client or in the communication channels between the client and server. Client-facing attacks on the server are less well investigated as the assumption is that a large collective of clients offer resilience. Here, for the first time, we show that by attacking clients that lead to repetitive clustered memory updates in the server, we can remote initiate a rowhammer attack to corrupt the server memory with induced bit-flips. A reinforcement learning (RL) attacker without backdoor access to the server can learn how to maximize server repetitive clustered memory updates by manipulating the clients' sensor observation. We demonstrate the feasibility of our attack using FL systems with sparse updates, tasked for automatic speech recognition (ASR) as well as CIFAR10 classification, our adversarial attacking agent can achieve around 70\% repeated update rate (RUR) in the targeted server model, effectively inducing bit flips on server DRAM. The security implications are that such adversarial observations can cause disruptions to learning or may inadvertently cause elevated privilege, which could pave the way for further research on practical mitigation strategies in FL and hardware design.
\end{abstract}


\section{Introduction}

\subsection{Motivation}

\begin{figure}[htbp]
    \centering
    \includegraphics[width=1\linewidth]{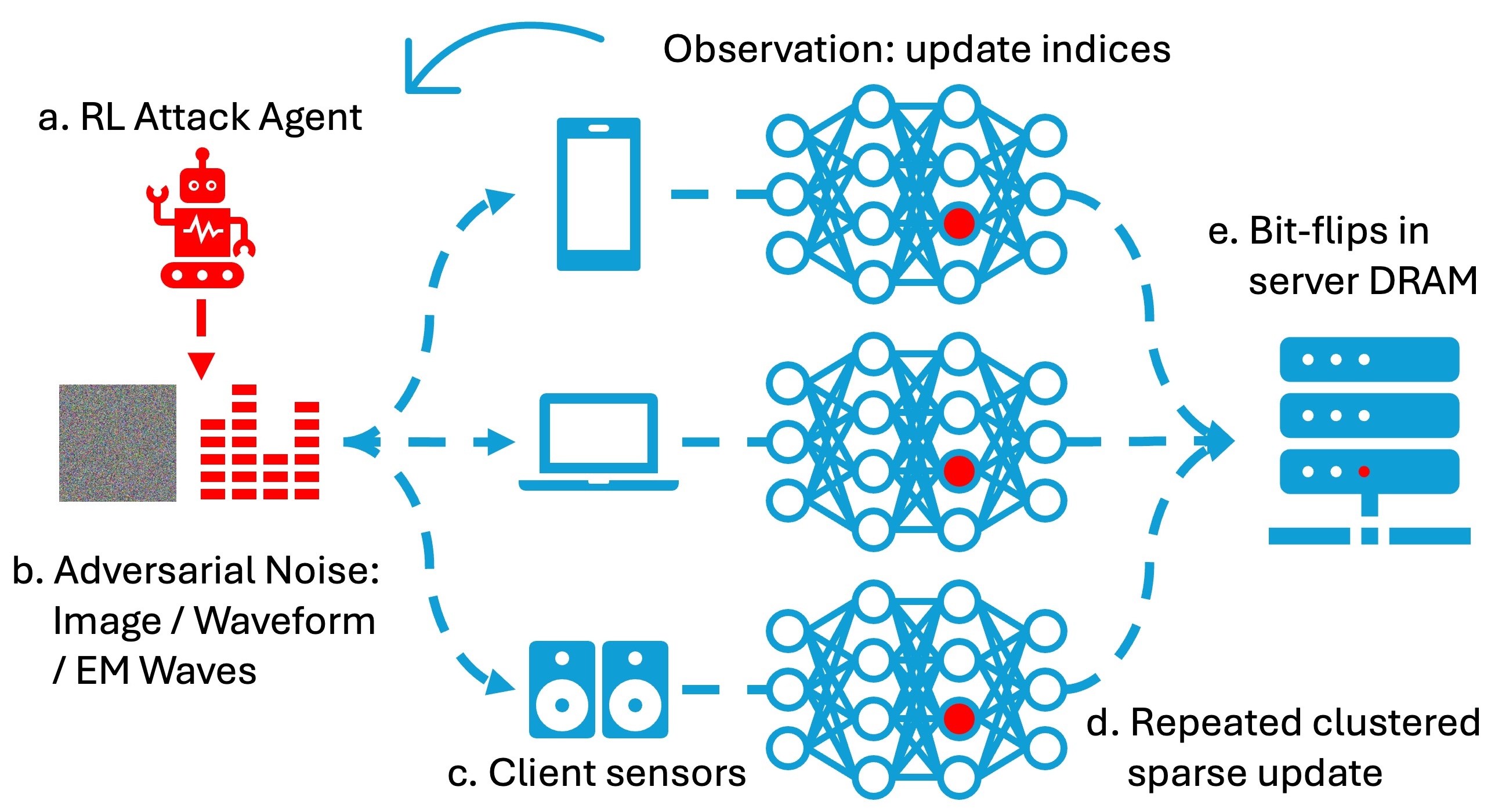}
    \caption{Framework of our proposed attack vector. a) a PPO agent generates b) adversarial noises in the physical domain in the form of image patterns or waveforms that interfere the inputs of c) client sensors of FL, resulting the clients to send d) repeated clustered updates, which could trigger e) Rowhammer attack (bit-flips) on server DRAM.}
    \label{fig:sys_diag}
\end{figure}

Recent advances in artificial intelligence (AI), particularly in federated learning (FL), have significantly enhanced privacy preservation by enabling distributed model training across numerous client devices without sharing raw data~\cite{mcmahan2017communication}. However, the increasing complexity and scale of these distributed AI systems also expose them to novel and sophisticated threat vectors that go beyond traditional adversarial machine learning techniques~\cite{kairouz2021advances}. Among hardware security attacks, the Rowhammer attack stands out for its ability to induce bit flips in Dynamic Random Access Memory (DRAM) through electromagnetic interference caused by frequent activations of adjacent rows~\cite{kim2014flipping}. Despite industry standard countermeasures such as Target Row Refresh (TRR), recent studies continue to show that modern DRAM modules remain vulnerable to advanced Rowhammer attacks~\cite{frigo2020trrespass, jattke2024zenhammer}.

Motivated by the following complementary observations, we explore the possibility of combining physical adversarial techniques with reinforcement learning (RL) methods to orchestrate Rowhammer bit flips remotely in FL environments. On the one hand, existing Rowhammer strategies typically assume some form of software level access to victim hardware~\cite{seaborn2015exploiting, gruss2016rowhammer, tatar2018throwhammer}. This assumption substantially limits their practical relevance in FL settings, where servers are usually difficult for potential aggressors to reach, especially at the hardware level. This challenge is expected to become even more pronounced in future deployments of FL over 6G radio access networks (RAN), where the server is likely to be placed within the central unit function and protected against backdoor entry or direct attacks. On the other hand, physical adversarial attacks have shown that carefully crafted electromagnetic or acoustic perturbations can mislead AI driven sensor inputs without direct digital intrusion~\cite{eykholt2018robust, carlini2018audio}.

Specifically, large scale FL systems often incorporate efficiency optimizations such as sparse gradient updates, pinned memory regions, huge pages, and Remote Direct Memory Access (RDMA), which may unintentionally increase susceptibility to carefully crafted physical perturbations. We hypothesize that these optimizations, while beneficial for performance, can facilitate indirect Rowhammer attacks by enabling finely controlled memory access patterns that are favorable for inducing DRAM bit flips. As a result, an adversary may exploit federated clients by perturbing their sensor inputs in the physical world with maliciously crafted signals, such as acoustic or electromagnetic noise, to indirectly influence server side memory access patterns and ultimately disrupt or degrade FL system performance without requiring direct access to the protected server hardware.

\subsection{Contributions}
In this paper, we propose a novel threat vector, see Fig.~\ref{fig:sys_diag}, that leverages a reinforcement learning-based Proximal Policy Optimization (PPO) agent to induce Rowhammer bit-flips on federated learning servers indirectly via physical interference at client devices. The PPO agent generates stealthy adversarial waveforms or image patches to be captured by client sensors (e.g. microphones, cameras), triggering clustered model parameter updates on the server. These updates, conveyed by efficiency optimizations such as RDMA, pinned memory, and sparse gradient updates, achieve the activation frequency and regularity necessary to induce Rowhammer attacks on server-side DRAM.

The key contributions of this work include:

\begin{enumerate}

\item \textbf{Novel Attack Vector}: We introduce the first physical-domain-driven remote Rowhammer attack vector specifically targets federated learning (FL) systems. Unlike traditional Rowhammer attacks, our approach does not require direct access or explicit control over victim hardware or software resources, greatly expanding its potential attack surface.

\item \textbf{PPO-based Attack Generation Framework}: We design a reinforcement learning framework that generates stealthy adversarial noises, targeting at causing repeated and clustered parameter updates within FL models, which consequently induce Rowhammer bit-flips. 

\item \textbf{Vulnerability Analysis}: We perform experimental evaluations using realistic federated learning scenarios with popular automatic speech recognition (ASR) models and computer vision models with common open datasets. Our experiments demonstrate that Rowhammer bit-flips can be induced on FL systems enhanced with common efficiency optimizations through adversarial interference on client sensors.

\item \textbf{Potential Mitigation Strategies}: We discuss practical countermeasures and security recommendations to defend against the proposed threat vector, such as adversarial input detection and systematic analysis of optimization technique deployment.

\end{enumerate}

\subsection{Organization}

The remainder of this paper is organized as follows. Section II provides a relevant background on DRAM architecture, Rowhammer vulnerabilities, FL security, and physical adversarial attacks. Section III formally describes the proposed attack vector, including motivation, feasibility analysis, and attack formulation. Section IV details the experimental setups, datasets, and evaluation. Section V presents experimental results, analyzing the effectiveness of the PPO-based adversarial attack framework. Section VI discusses the practical implications and limitations of our approach, mitigation techniques and recommendations to FL system design, as well as outlining directions for future research. Finally, Section VII concludes the paper.

\section{Background and Related Work}

\subsection{DRAM Architecture and Rowhammer Attack}
\begin{figure*}[htbp]
\centering
\includegraphics[width=0.98\textwidth]{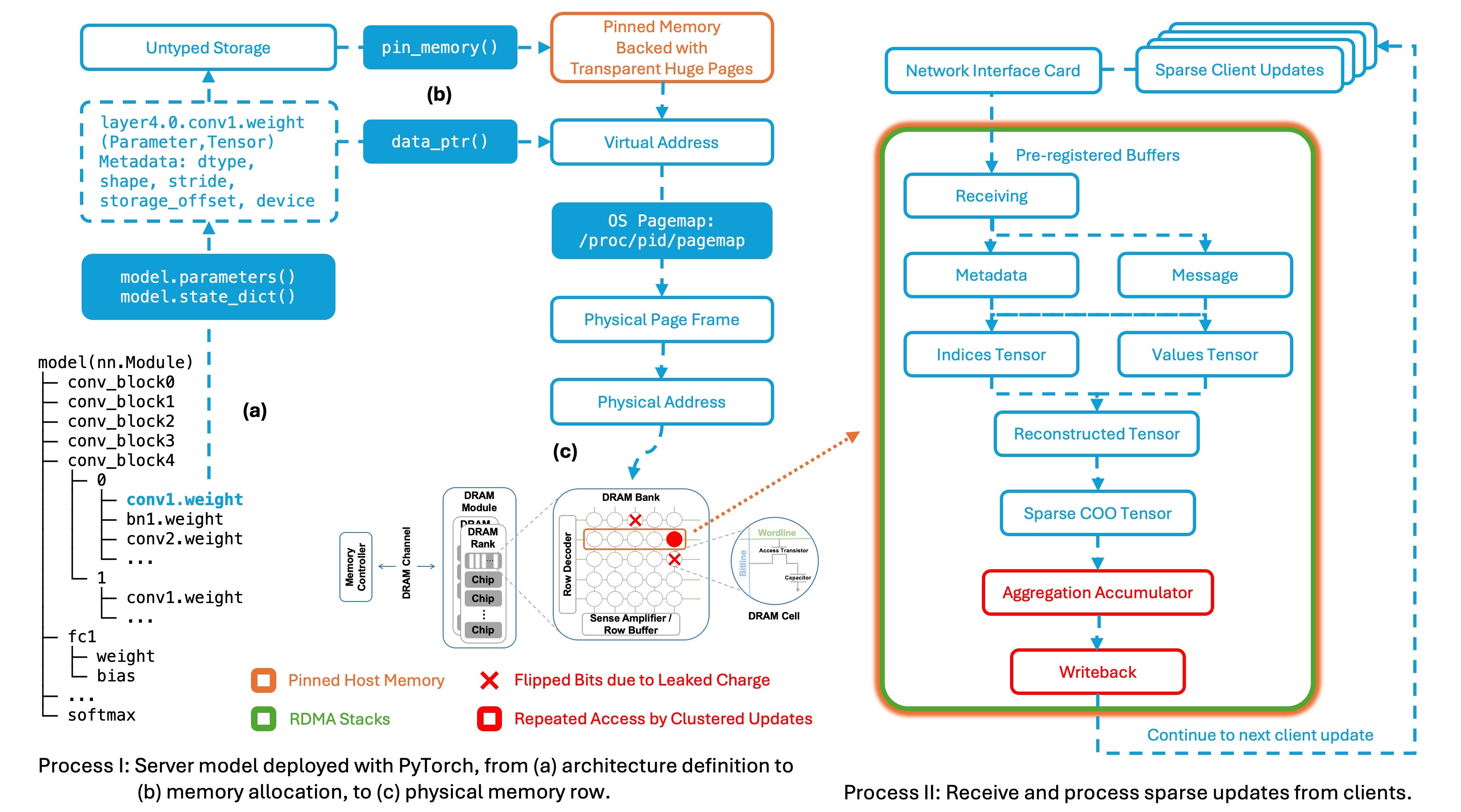}
\caption{Memory allocation of FL CNN server model with PyTorch (Process-I), and access pattern at communication when receiving and processing sparse updates from clients via RDMA (Process-II). I. the CNN model consists of a hierarchical python modules, taking the 'conv 1' layer from 'conv block 4' as example, the tensor is broken down into metadata and values before put into untyped storage in allocated memory, which is set to page-lock backed with Transparent Huge Pages, and mapped to a physical address of a few rows. II. When the server receive a queue of sparse updates from clients through RDMA, each message is broken down and reconstructed into a COO Tensor to be processed through the aggregation accumulators and writeback buffers of each layer tensor of the server model, which in the scenario of clustered sparse updates, these buffers (marked in red) are repeatedly accessed, serving as hammer row, potentially causing bit-flips in adjacent rows}
\label{fig:organization}
\end{figure*}

Modern DRAM modules can be classified into single-rank, dual-rank, and multiple-rank according to the number of ranks. Each rank includes multiple chips that share the same data bus. Each chip contains multiple banks, which are organized into bank groups. It is necessary to identify the bank group number when accessing one specific bank. A DRAM bank consists of an array of memory cells, where all cells in a horizontal direction are collectively referred to a DRAM row, while those in a vertical direction are referred to a DRAM column. Each cell represents one logical bit. Depending on the memory strategy, a fully charged cell indicating a logical '1' is called a true-cell, whereas its counterpart, the anti-cell, denotes a logical '0'. The specific DRAM organization is shown in Fig.~\ref{fig:organization}. When data needs to be read from or written to DRAM, the memory controller (MC) issues an \texttt{ACT} command to open a row. The entire row of data is activated by the wordline and transferred to the sense amplifiers (also known as the row buffer). The bitlines then access specific columns within the row buffer to perform the read or write operation. Finally, the MC issues a \texttt{PRE} command to precharge the bitlines and close the currently active row.

For DRAM access, whether it is a read or write operation, the MC issues both an \texttt{ACT} command and a \texttt{PRE} command for the accessed row. The fundamental cause of the Rowhammer attack lies in the frequent issuance of \texttt{ACT} commands. When the adjacent rows (aggressor rows) of a victim row are repeatedly activated, the wordline voltage, which drives the aggressor row, changes rapidly and frequently. It then induces electromagnetic coupling with the memory cells in the victim row, altering their charge state. It should be noted that even if only a single aggressor row is used for the Rowhammer attack, it still requires the coordinated activation of two rows within the same bank. This is because if only one aggressor row is accessed repeatedly, the MC will not precharge this row before completing all operations, making the attack ineffective.

Furthermore, the reason why the Rowhammer attack requires rapid and frequent access to the aggressor row lies in the DRAM refresh mechanism. DRAM performs periodic refresh operations to maintain the charge state of all memory cells. Typically, this refresh period is $64~\text{ms}$ or $128~\text{ms}$, but under high-temperature conditions, it could be reduced to $32~\text{ms}$. In each refresh period, the MC typically issues 8192 \texttt{REF} commands, which sequentially refresh all rows on the DRAM module. Taking a $64~\text{ms}$ refresh period as an example, the average interval between each \texttt{REF} command is $7.8~\mu\text{s}$. If the number of activations on the aggressor row fails to effectively cause charge leakage in the victim row before its next refresh, the attacker must accumulate more activations again in next refresh period to execute Rowhammer attack.

The factors influencing the Rowhammer attack include manufacturing, data pattern, and access pattern. Limited by the current manufacturing process, there are some inherently vulnerable memory cells in each DRAM chips that are more prone to charge leakage compared to normal cells, regardless of other factors. The data pattern refers to the scenario where the Rowhammer attack is more likely to succeed when the charge states of memory cell in the aggressor row and victim row are different~\cite{mutlu2019rowhammer}. Although adversary attempting to tailor the data in aggressor row based on the victim data may be affected by anti-cells, existing research has proved that the proportion of anti-cells with the same chip is very small~\cite{kim2014flipping}. The access pattern includes single-sided, double-sided, and many-rows Rowhammer attack~\cite{frigo2020trrespass, hassan2021uncovering}. The double-sided Rowhammer, compared to a single-sided, involves two aggressor rows targeting the same victim row. This type of attack is generally more effective than single-sided Rowhammer. However, it requires more detailed memory mapping preparation. On the other hand, the many-rows Rowhammer is often used to bypass the TRR protection mechanism.

The key points and progress details of executing and exploiting a Rowhammer attack are as follows:
\begin{enumerate}
    \item \textbf{Selecting Rowhammer access patterns:} The specific DRAM modules of target system guides the choice of Rowhammer attack access patterns, primarily depending on whether the victim chips have TRR enabled. For early DDR3, both simple single-sided and more effective double-sided Rowhammer are ideal access patterns. However, for modern DDR4 chips, where TRR is widely implemented, many-rows Rowhammer and its variants must be used instead~\cite{frigo2020trrespass,de2021smash}. This is because TRR protection mechanism rely on its internal sampler~\cite{jiang2021trrscope}. By using multiple dummy rows, attackers can successfully overload the size of sampler, preventing the actual aggressor rows from being affected by extra refresh operations. It is worth noting that error-correcting code (ECC) has been proven ineffective against this attack~\cite{aweke2016anvil}, as efficient access patterns such as double-sided Rowhammer can also overwhelm ECC mechanisms~\cite{cojocar2019exploiting}.
    
    \item \textbf{Determining Address Mapping:} Since Rowhammer attack is triggered by adjacent rows of victim data, determining the addresses of aggressor rows is crucial. However, the virtual memory address used in application do not correspond to actual DRAM address. The process of locating the correct address typically involves two mapping stages~\cite{cojocar2020we}. The first is application-level virtual address to physical address. In operating system, the memory management unit (MMU) in the CPU, maps contiguous virtual addresses to non-contiguous physical addresses in a non-linear manner. In some operating systems, such as Ubuntu, the mapping can be retrieved via \texttt{/proc/pid/pagemap}. The next stage involves physical address to DRAM address mapping, which varies by CPU architecture and DRAM manufactory. This mapping is undocumented, and common approaches to reverse-engineering it rely on side-channel analysis~\cite{jattke2024zenhammer, xiao2016one, kang2024sledgehammer, pessl2016drama}. A key point to mention is that in a double-sided Rowhammer scenario, a common strategy for finding usable contiguous physical addresses is to use huge pages.

    \item \textbf{Avoiding Cache Effects:} Frequently accessed data will be stored in the cache to avoid CPU complex operations. Thus, Rowhammer attack requires to use specialized CPU instructions to bypass cache to achieve high-frequency activations of aggressor rows in real attack scenarios. Currently, the main instructions used to achieve this purpose include \texttt{clflush}~\cite{qiao2016new, kim2014flipping}, \texttt{clflushopt}~\cite{jang2017sgx}, and \texttt{invd} or \texttt{wbinvd}\cite{cojocar2020we}. In a carefully crafted attack sequence, incorporating the aforementioned instructions can help the attacker bypass the cache. Sometimes, this also requires the use of memory barriers such as \texttt{mfence}, \texttt{lfence}, and \texttt{sfence}, but they may not achieve the optimal activation rate. Listing~\ref{code:rowhammer} illustrates a Rowhammer attack sequence utilising the \texttt{clflushopt} instruction and memory barrier. Additionally, there are some attacks based on cache line conflicts to bypass ~\cite{aweke2016anvil, lipp2020nethammer}. 

    \item \textbf{Ensuring Effective Activation Rate and Count:} After determine all above conditions, attackers should ensure enough activations to induce bit-flips. There are two key factors to ensure the successful execution of the attack. One is maintaining an effective activation rate. Even though the standard tRC (Row Cycle Time) is only $46~\text{ns}$, the average interval between two activation instructions should be around $220~\text{ns}$, considering the influence of other instructions. The other factor is accumulating a sufficient number of activations to ensure that the aggressor rows are activated enough times to induce bit flips. Additionally, when TRR are present, the dummy rows used to bypass TRR must also be accounted for in the total activation count. The total number of activations should not exceed the maximum allowable activations within a single refresh period ($64~\text{ms}$) to avoid DRAM refresh interventions.
\end{enumerate}

\begin{center}
\begin{lstlisting}[caption={Pseudo code of Rowhammer attack utilising the \texttt{clflushopt} instruction}, label={code:rowhammer}]
        loop:
            mov (X), %eax
            mov (Y), %ebx
            clflushopt (X)
            clflushopt (Y)
            mfence
            jmp loop
            
\end{lstlisting}
\end{center}

\subsection{Federated Learning} 

Federated Learning (FL) is a distributed machine learning paradigm that enables collaborative model training across multiple devices while retaining the raw data locally~\cite{mcmahan2017communication}. Unlike conventional centralized training schemes, FL enhances data privacy by transmitting only model updates, e.g., gradients, to a central server for aggregation. This distributed framework is particularly advantageous for use cases where data confidentiality and regulatory compliance are paramount (e.g., healthcare, finance, or mobile devices). Moreover, the scalability of FL allows diverse and expansive datasets to be leveraged by increasing the number of participating clients, thereby improving generalization across heterogeneous data distributions. The typical FL process involves the following steps:

\begin{enumerate}
    \item \textbf{Local Training:} Each client trains a local model using its own private dataset.

    \item \textbf{Model Aggregation:} In each communication round, a designated subset of clients transmits model updates (e.g., weights or gradients) to the central server. These updates are then aggregated into a global model using algorithms such as FedAvg~\cite{mcmahan2017communication}.

    \item \textbf{Redistribution:} The aggregated global model, depending on the training strategy, is redistributed to the participating clients, or all clients, which then continue local training in the next round.
\end{enumerate}

\subsubsection{Security Threats in Federated Learning}

Federated learning (FL) systems are vulnerable to a variety of security threats that arise from both adversarial machine learning (ML) techniques and conventional communication/network-based attacks. The former category includes model poisoning, backdoor, and inference attacks, while the latter mainly consists of eavesdropping and distributed denial-of-service (DDoS) attacks. In this paper, we focus on an attack vector more closely aligned with adversarial ML methodologies.

\begin{itemize}
    \item \textbf{Poisoning Attacks:}
    Poisoning attacks are a specialized form of injection attack that targets ML models. Within FL, adversaries exploit malicious or compromised clients to manipulate the training process, which can degrade the global model's performance. Specifically, \textit{data poisoning} involves injecting or modifying training samples (often through mislabeled or corrupted data), whereas \textit{model poisoning} targets model updates (e.g., gradients or weights) to undermine accuracy or embed hidden behaviors.

    \item \textbf{Backdoor Attacks:}
    Backdoor attacks implant triggers in an ML model during training, allowing the adversary to covertly control the model output. By submitting local updates with deliberately engineered triggers or patterns, malicious clients can force the global model to behave normally for most inputs, yet generate attacker-controlled outputs for specific trigger inputs.
    These attacks are especially dangerous because they usually go undetected until the trigger is activated.

    \item \textbf{Inference Attacks:}
    Inference attacks exploit partial information shared in FL systems to reveal or deduce private data from participating clients.
    By analyzing the distributed global model, adversaries may ascertain membership (i.e., whether a particular sample was part of the training set) or other sensitive attributes.
    This highlights the necessity for more robust privacy-preserving techniques in FL.
\end{itemize}

These threats are traditionally studied at the machine learning level, where the goal is to compromise model utility, implant hidden behavior, or extract private information. Our perspective is slightly different. We consider whether the same attacker controlled update channel can also be used to influence low level system behavior, particularly memory access patterns on the server, thereby creating conditions that are relevant to Rowhammer style attacks.

\subsubsection{Efficiency Optimization}

Federated learning systems deployed in high performance computing facilities and data centres often adopt aggressive communication and memory optimizations to reduce training overhead. Among the most relevant are sparse updates, page locked memory, huge pages, and DMA or RDMA based data transfer. While these mechanisms are introduced for performance reasons, they can also shape memory access behavior in ways that become security relevant.

\begin{itemize}
    \item \textbf{Sparse updates} reduce communication overhead by transmitting only a subset of model parameter changes, such as those whose magnitudes exceed a threshold or rank among the largest entries. In a typical federated learning workflow, each client computes local gradients and sends only the selected updates to the server, which then aggregates these sparse contributions into the global model. This design can substantially reduce bandwidth consumption and accelerate training, especially when the federation contains a large number of clients. At the same time, repeatedly focusing communication on a small subset of parameters can make memory access patterns more concentrated and predictable, which is potentially relevant from an attack perspective.

    \item \textbf{Page-lock memory}, also referred to as pinned memory \cite{cuda_guide}, allows application-allocated virtual memory to be fixed to physical pages, preventing it from being swapped out or migrated by the operating system and thus avoiding additional memory remapping overhead. It is commonly used in DMA/RDMA operations and high-speed data transfers in GPU-based systems. In federated learning, pinned memory is typically adopted on the server side to improve the efficiency of receiving model parameters transmitted from clients.
    From a system security perspective, however, during the pinned period, the parameters transmitted by clients may repeatedly access a fixed physical memory region on the server, as illustrated in Fig.~\ref{fig:organization} (a) and (b). This behavior provides a potential opportunity for the frequent activation of specific physical rows, which is a necessary condition for enabling Rowhammer attacks.

    \item \textbf{Huge Pages} are a memory management mechanism that enables the operating system to use larger page sizes than the default 4KB (e.g., 2MB), thereby reducing the number of page table entries, lowering address translation overhead, and improving Translation Lookaside Buffer (TLB) efficiency \cite{linux_thp}. This technique is commonly adopted in high-performance computing (HPC) and memory-intensive applications to enhance overall memory access performance. In federated learning scenarios, where model parameters and intermediate data can be large, the server may employ huge pages to improve the efficiency of parameter buffer accesses.
    From an attacker's perspective, however, the use of huge pages leads to more contiguous physical memory allocation, which increases the likelihood that different memory access regions exhibit spatial locality in the underlying DRAM. This, in turn, may increase the probability that adjacent physical rows are involved in memory accesses, thereby providing a potential basis for satisfying the physical adjacency condition required by Rowhammer attacks.

    \item \textbf{DMA and RDMA} Direct Memory Access (DMA) and Remote Direct Memory Access (RDMA) are widely adopted high-performance data transfer mechanisms in modern computing systems. DMA allows devices to access main memory directly without CPU involvement, while RDMA further enables memory-to-memory data transfer across networked nodes, significantly reducing latency and improving bandwidth utilization. RDMA has been extensively used in high-performance computing (HPC) and data centre environments to accelerate distributed machine learning training, particularly for efficient parameter synchronization and system scalability \cite{zhang2024fedrdma}. In practice, RDMA can achieve communication latencies as low as 1-$2\,\mu\text{s}$ and support bandwidths exceeding 100 Gbps and even up to 200 Gbps, making it well-suited for federated learning systems to reduce communication overhead between clients and servers and improve overall training efficiency.
    From a memory access perspective, DMA/RDMA alters the traditional data path that relies on CPU cache hierarchies, allowing data transfers to interact more directly with main memory. During continuous parameter uploads from clients, this mechanism enables the server-side memory buffer to experience more intensive and stable write operations. From an attacker's perspective, the ability to bypass CPU caches provides a direct access path to DRAM, which is a necessary condition for Rowhammer attacks. Moreover, the combination of high throughput and low latency may further increase the access intensity to specific memory regions, potentially amplifying memory disturbance effects.
\end{itemize}

\subsection{Physical Adversarial Attacks}

Deep learning systems have been challenged across a wide range of tasks, including computer vision, speech processing, autonomous driving, surveillance, biometric authentication, robotics, and voice controlled services, by adversarial attacks, in which carefully crafted perturbations cause a model to produce incorrect predictions. While early studies focused on perturbations applied directly in the digital domain, later work showed that adversarial manipulations can also remain effective after physical world transformations~\cite{szegedy2013intriguing,goodfellow2014explaining,kurakin2018adversarial}.

In computer vision, adversarial perturbations have been demonstrated on printed images, adversarial patches, wearable objects, and modified traffic signs, showing that malicious inputs can preserve their effect even under viewpoint changes, lighting variation, and environmental noise~\cite{eykholt2018robust,brown2017adversarial,evtimov2017robust,athalye2018synthesizing,duan2020adversarial}. In the audio domain, adversarial perturbations can be embedded into speech or music so that automatic speech recognition systems produce attacker chosen transcriptions while the signal remains natural or inconspicuous to human listeners~\cite{carlini2018audio,yuan2018commandersong,zhou2019hidden}. These results show that adversarial manipulation is not limited to synthetic digital settings, but can be embedded into real world inputs that are naturally collected by machine learning systems.

This observation is relevant to federated learning because client side training data may originate from physical environments, such as cameras, microphones, wearable sensors, or mobile devices. An adversary who can influence these inputs may be able to steer local model updates without needing direct access to the global model or server side infrastructure. In this sense, physical adversarial examples provide a practical threat vehicle for shaping the behavior of FL clients through apparently legitimate data without tackling with authentication processes.

In this paper, however, our focus is not on the misclassification effect of physical adversarial examples themselves. Instead, we use them to motivate a broader point: attacker controlled inputs and updates in FL can influence not only model behavior, but also the system level processing of selected parameters. This perspective is important for the attack vector developed next, where repeated and targeted updates interact with efficiency optimizations to create conditions favorable to Rowhammer style exploitation.

\begin{figure}[htbp]
    \centering
    \includegraphics[width=1\linewidth]{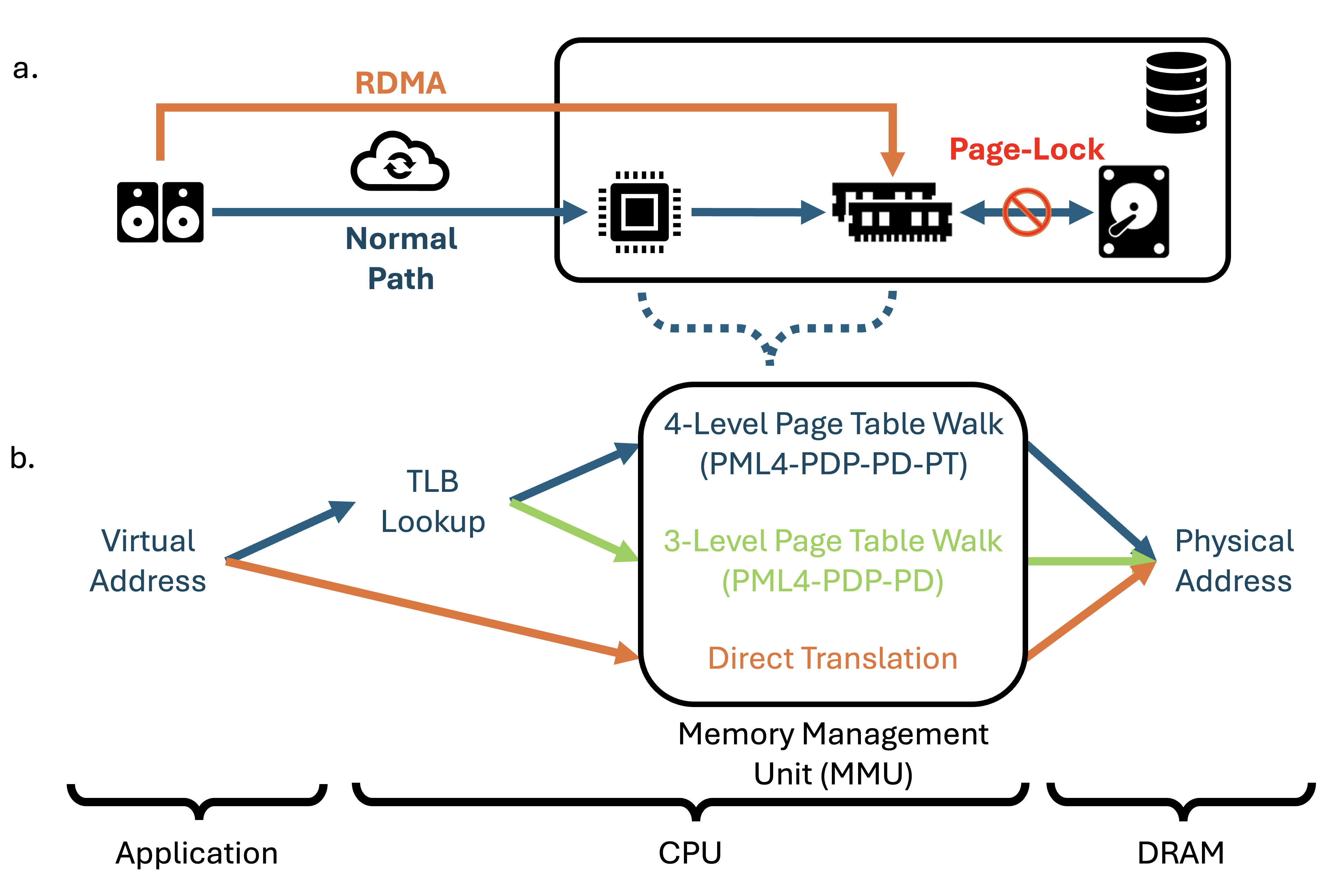}
    \caption{DRAM access pathways and processes of FL. a) DRAM access pathways. For normal access, client updates go through network to the CPU of server before getting to the DRAM of the server, while RDMA allows clients to bypass the CPU; b) DRAM access processes. Normally, when mapping virtual addresses of applications to physical addresses on DRAM, the CPU lookup TLB, and walk through the page table if TLB miss. Using \textcolor{green}{Huge Page} reduces a level in page table walk, and \textcolor{orange}{DMA/RDMA} can bypass both TLB lookup and page table walk. \textcolor{red}{Page-lock} prevents variables from offloading from DRAM to disk.}
    \label{fig:va2pa}
\end{figure}

\section{Methodology}

\subsection{Attack Vector Formulation}

\subsubsection{Predictable Memory Access Patterns}

In high-performance computing (HPC) environments, optimizations meant to enhance efficiency can inadvertently introduce vulnerabilities exploitable by Rowhammer-style attacks. We identify three specific properties induced by optimizations, as shown in Fig.~\ref{fig:va2pa} commonly found in federated learning (FL) systems, particularly suited for exploitation:

\begin{itemize}

    \item \textbf{Predictable and stable physical placement.}  
    Modern federated learning systems frequently use pinned huge pages to accelerate data movement between CPUs and GPUs. Huge pages reduce translation overhead, and page locking prevents the allocated region from being swapped out or migrated. DMA and RDMA then provide direct access to these pinned buffers. Together, these mechanisms create memory regions that are both physically contiguous and temporally stable, especially within the short time scale relevant to Rowhammer, such as a 64 ms DRAM refresh window. This stability can significantly reduce the uncertainty that attackers usually face when attempting to align accesses with a chosen physical DRAM row. In effect, the federated learning workflow itself may help an attacker maintain a target row at a fixed physical location by repeatedly interacting with selected model parameters.

    \item \textbf{Direct DRAM access with reduced cache disruption.}  
    Efficient distributed training often depends on RDMA, DMA, sparse synchronization, and non temporal accesses. These features are beneficial for throughput, but they also reduce the degree to which CPU caches obscure or absorb repeated memory references. RDMA zero copy transfers allow remote actors to trigger frequent reads and writes directly against the target node's memory, while sparse updates repeatedly touch only a narrow set of parameters. This combination provides a practical way to generate focused and repeated accesses to the same pinned memory regions, thereby increasing the likelihood of direct DRAM row activations. Non temporal instructions strengthen this effect by further suppressing cache residency.

    \item \textbf{High frequency and persistent hammering opportunities.}  
    Successful Rowhammer attacks require a sufficiently high rate of repeated row activations. Federated learning optimizations can make such access patterns easier to sustain. Reduced latency, lower jitter, and concentrated communication around sparse gradients allow attackers to issue highly repetitive memory operations with less variability. This effect becomes even stronger in deployments with many clients, where repeated sparse contributions can collectively amplify hammering intensity. Gather scatter aggregation, especially when restricted to the most significant gradients such as the top 0.05\%, repeatedly targets the same small parameter subset. From an attack perspective, this creates a natural mechanism for persistent hammering of memory regions associated with those parameters.
\end{itemize}

\subsubsection{Stealthy Attack Vehicle}

In typical federated learning deployments, corrupting the authentication process or injecting malicious clients at scale is challenging. Yet physical interferences via adversarial noises are often overlooked. Rather than requiring sophisticated software exploits or the circumvention of cryptographic authentication protocols, an attacker only needs to alter the physical space with adversarial patches within a camera's field of view, or by broadcasting imperceptible, high-frequency audio perturbations in the vicinity of a voice-controlled IoT device. When the FL client samples its environment, this physical adversarial noise translates into the client's local parameter updates, infiltrate the authentication process and alter server memory access patterns.

\subsubsection{Reinforcement Learning Formulation}

We initiate a feasibility validation of the attack vector with a reinforcement learning agent targeting a FL system using page-lock memory backed with Transparent Huge Page, and send sparse updates to the server via RDMA. The agent can observes the inputs and parameter updates of the clients, and generates adversarial noises that feeds to the clients' input after emulation of physical embodiment and re-digitalization.

\begin{table*}[!t]
    \caption{Target Models with Sizes and Computation Precision.}
    \label{tab:models}
    \centering
    \begin{tabular}{lccccc}
        \hline
        \textbf{Model } & \textbf{Params(M)} & \textbf{Trainable Tensors} & \textbf{Precision} & \textbf{$H_{\text{max}}(0.1\%)$} & \textbf{$H_{\text{max}}(0.05\%)$}\\
        \hline
        Conformer-CTC-S~\cite{gulati2020conformer}     & 8.7  & 480 & \multirow{2}{*}{INT4} & 296K & 592K\\
        Squeezeformer-XS ~\cite{kim2022squeezeformer}  & 9.0  & 480 & & 286K & 572K\\
        QuartzNet 5x5~\cite{kriman2020quartznet}       & 6.7  & 130 & \multirow{2}{*}{INT8} & 192K & 384K\\
        MobileNet V3 Small ~\cite{howard2019searching} & 2.9  & 142 & & 444K & 888K\\
        \hline
    \end{tabular}
\end{table*}

\subsection{Evaluation Metrics}

\subsubsection{Notation}

To evaluate the proposed attack strategy, we relate repeated sparse parameter updates in federated learning to the number of effective DRAM row activations that may be induced within a refresh window. We use the following notation:

\begin{itemize}
    \item $p$: Fraction of model parameters included in each sparse update.
    \item $S_{\text{update}}(p)$: Size in bytes of one sparse update when the update fraction is $p$, including both parameter values and indexing metadata.
    \item $H_{\max}(p)$: Upper bound on the number of sparse updates that can be processed within one DRAM refresh period for update fraction $p$.
    \item $\Delta t_{\text{refresh}}$: DRAM refresh period.
    \item $BW$: Effective bandwidth available to sparse update processing.
    \item $T_{\text{flip}}(r)$: Rowhammer threshold of row $r$, measured as the number of effective row activations required to induce a bit flip.
    \item $\overline{T_{\text{flip}}}$: Average Rowhammer threshold across tested vulnerable rows.
\end{itemize}

\subsubsection{RL Performance Metrics}

As illustrated in Process-II in Fig~\ref{fig:organization}, centred to the success of our attack vector is inducing consistent repeated and clustered sparse updates from the clients. For the RL agent, we use metrics that quantify the persistence and spatial concentration of the induced sparse updates.

\begin{itemize}
    \item \textbf{Repeated Update Rate (RUR):}
    RUR measures the overlap between updated parameter sets in consecutive episodes or rounds. A higher RUR indicates that the agent repeatedly targets similar parameter indices:
    \begin{equation}
        \mathrm{RUR}
        =
        \frac{\sum_{t=1}^{T-1} \lvert U_t \cap U_{t+1} \rvert}
             {\sum_{t=1}^{T-1} \lvert U_t \rvert},
    \end{equation}
    where $U_t$ denotes the set of parameter indices updated at step $t$.

    \item \textbf{Cluster Density (CD):}
    CD measures how tightly updated indices are concentrated in the global parameter index space. Lower CD indicates stronger spatial locality:
    \begin{equation}
        \mathrm{CD}
        =
        \frac{L}{\lvert \boldsymbol{\theta} \rvert},
    \end{equation}
    where $\boldsymbol{\theta}$ is the full parameter vector, $k$ is the number of updated indices in one step, $m = \lceil 0.9k \rceil$, $i_{(1)} \leq i_{(2)} \leq \dots \leq i_{(k)}$ are the sorted updated indices, and
    \begin{equation}
        L
        =
        \min_{1 \leq j \leq k-m+1}
        \left(i_{(j+m-1)} - i_{(j)} + 1\right)
    \end{equation}
    is the minimum span containing 90\% of updated indices.
\end{itemize}

Together, RUR and CD characterize whether the agent repeatedly concentrates updates on a small and stable region of the model, which is a necessary precursor to repeated accesses to the same memory region.

\subsubsection{Bit Flip Validation}

Given the effective bandwidth $BW$, refresh period $\Delta t_{\text{refresh}}$, and sparse update size $S_{\text{update}}(p)$, the maximum number of sparse updates that can be processed within one refresh window is upper bounded by
\begin{equation}
    H_{\max}(p)
    =
    \frac{BW \cdot \Delta t_{\text{refresh}}}{S_{\text{update}}(p)}.
\end{equation}

This quantity is an upper bound because it does not account for protocol overhead, software processing cost, cache effects, or other system bottlenecks.

To estimate the number of effective activations on a targeted DRAM row, we combine the maximum update rate with the measured repeated update behavior. Then the expected number of effective activations within one refresh window is approximated as
\begin{equation}
    \mathbb{E}[A]
    =
    \mathrm{RUR} \cdot H_{\max}(p).
\end{equation}

Finally, for a given update fraction $p$, the attack is considered capable of creating Rowhammer relevant conditions when
\begin{equation}
    \mathbb{E}[A] \geq T_{\text{flip}}(r)
\end{equation}
for a targeted vulnerable row $r$, or more conservatively when $\mathbb{E}[A]$ approaches empirically observed Rowhammer thresholds on the test platform. In practice, this comparison should be interpreted as a feasibility criterion rather than a deterministic guarantee of bit flips.

\subsection{Target System}

The target systems consist of two variant, a self-supervised federated voice recognition system, and a federated CIFAR10 classifier, emulating large scale IoT or voice assistant pre-training systems, which are widely deployed on various platforms including smart home devices, mobile phones, and other edge devices, contributing to a large quantity of clients.

\subsubsection{Model Structure}

Model structures of the target systems, as well as details including the corresponding sizes and precision of these models are presented in Table~\ref{tab:models}.

\subsubsection{Dataset}

For dataset, the ASR systems take the Common Voice 17.0 dataset~\cite{commonvoice2020} as the training data, which is a large-scale multilingual voice dataset. We only use the English subset for simplicity, which contains 1.1M voice clips with a total duration of more than 3k hours. As for the computer vision target system uses the CIFAR10 dataset~\cite{krizhevsky2009learning}. In both cases, the data is randomly split into equal client partitions, where the size of the partition is determined by the number of clients.

\subsubsection{Server Setups}

On the server side, global model and aggregation are performed using huge page and page-locked memory, and the communication between the server and clients are optimized using RDMA.

\subsubsection{Clients Setups}

On the client side, for the ASR systems, each client emulates different physical environments via applying random Gaussian noise in addition to the adversarial noise generated by the RL agent, after which the mixed waveforms are resampled to 16kHz to match the Common Voice dataset. For computer vision, the printing and re-digitalization of the dataset is emulated with a process including paper texture masking, ink bleeding, gamma and rescaling. Also, the clients are set to send sparse updates of the most significant gradient updates to the server. The total number clients do not directly affect the attack vector, see Process-II in Fig.~\ref{fig:organization}, but indirectly through update scheduling.

\subsection{RL Agent Design}
\label{subsec:rl_design}
We propose a PPO-based agent to generate adversarial waveforms or image perturbations that consistently induce parameter updates clustered within targeted indices of the target models' parameter vector, denoted by $\boldsymbol{\theta} \in \mathbb{R}^{M}$. The agent learns under the guidance of client parameter updates.

\subsubsection{\textbf{Observation and Action Spaces}} 
\label{subsec:obs_action}

In \textbf{Stage 1}, observations explicitly encode parameter update locations $\mathbf{u}_t$ to facilitate clear mapping from adversarial perturbations to model updates in initial learning, aligning with curriculum-learning principles~\cite{bengio2009curriculum}:

\begin{itemize}
    \item \textbf{Observation:} $\mathcal{O} = \{x_{\text{clean}}, \mathbf{u}_t\}$, with binary vector $\mathbf{u}_t \in \{0,1\}^{M}$ indicating recently updated parameters.
    \item \textbf{Action:} $\mathcal{A} = \{\delta \in \mathbb{R}^D \mid \|\delta\|_\infty \leq \epsilon\}$, constrained adversarial perturbations. To improve learning efficiency, action space of the agent is compressed with a patch encoder. The input perturbations are parameterized in a low-dimensional latent patch space. For ASR systems, the patches are a set of equal-length waveforms, and for the computer vision system, the patches are grids of sizes such as 8x8, 16x16. Then, a fixed decoder maps these patches to full-resolution perturbations, reducing dimensionality while preserving spatial structure. This latent representation serves as a unified action space for reinforcement learning.
\end{itemize}

\subsubsection{\textbf{Reward Formulation}}
\label{subsec:reward}

Both stages share a unified reward structure composed of three components balancing localization, clustering specificity, and imperceptibility constraints:

\begin{equation}
    R_t = \underbrace{\alpha\,\text{EMD}(\mathbf{u}_t,\mathbf{u}_{t-1})}_{\text{Update Stability}} + \underbrace{\beta\,\frac{\|\mathbf{u}_t[i:j]\|_0}{\|\mathbf{u}_t\|_0}}_{\text{Target Focus}} - \underbrace{\gamma\mathcal{L}_{\text{perc}}}_{\text{Stealth}}
\end{equation}

\textit{Update Proximity}: Earth Mover's Distance (EMD) penalizes large shifts between consecutive update indices, promoting stable parameter clustering.

\textit{Target Specificity}: Ratio term emphasizes updates within targeted parameter indices $[i:j]$, here, the target is acquired not by designation, but rather condensed from observed preliminary dominant indices in early rounds.

\textit{Perceptibility Penalty}: 
For ASR targets, the perceptual constraint simultaneously limits frequency-domain artifacts and temporal waveform distortions~\cite{carlini2018audio}, calculated as follows.
\begin{equation}
    \mathcal{L}_{\text{perc}} = \lambda_1\|\text{STFT}(\mathcal{A} + x_{\text{clean}}) - \text{STFT}(x_{\text{clean}})\|_F + \lambda_2\,\text{rms}(\mathcal{A})
\end{equation}
where STFT is the Short-Time Fourier Transform, rms is root mean square, and $\lambda_{1,2}$ are weight factors.

For CV targets, the constraint is $\mathcal{L}_{\text{perc}} = \lambda\|\mathcal{A}\|_2^2$.

\subsubsection{\textbf{Policy Network Architecture}} 
\label{subsec:policy_net}

The policy network consists of a target specific feature extractor and multi-layer perceptron decoders. The feature extractor shares the typical structures of the model of the target system. For the ASR targets, the extractor first processes raw waveforms through a multi-resolution convolutional frontend, followed by bidirectional temporal modeling, before diverging into stage-dependent perturbation generators, while for the computer vision target, the extractor consists of convolution blocks like that of a ResNet model~\cite{he2016deep}. 

\section{Experiments}

\subsection{Experimental Setups}

This work utilizes the DRAM Bender open-source platform to test the Rowhammer attack threshold~\cite{olgun2023dram}. The platform is built on the Alveo U200 Accelerated Card~\cite{alveo} and is designed to emulate a memory controller using an FPGA, allowing us to precisely issue Rowhammer-related instructions. The specific memory module tested is the MTA18ASF2G72PZ-2G3B1-16GB~\cite{memory}, which complies with the standard DDR4 specification. Additionally, it has data rate of 2400 MT/s and bit-width of 72 bits. After excluding 8 bits of ECC, the effective bit-width for transmission is 64 bits:

\begin{equation}
\label{eq:bandwidth}
\text{$BW$} = \frac{\text{Data Rate} \times \text{Bit-Width}}{8 \text{ bits/byte}}
\end{equation}

According to equation \eqref{eq:bandwidth}, we can first get the bandwidth of tested module is 18.75 GB/s. Therefore, the data throughput for each 64ms refresh period will be approximately 1.2 GB.
Furthermore, the default \textbf{tRC} is $46.16~\text{ns}$. 

In an operating system-level attack environment, the optimal \texttt{ACT} rate based on the \texttt{clflushopt} instruction sequence is 159 ACTs/tREFI, meaning that the actual average \textbf{tRC} used for frequent activations is $49~\text{ns}$, which is close to the theoretical ACT limit. Under this condition, the attacker can issue up to 1,306K activations per refresh window. Moreover, all other timing parameters of tested module follow the default configuration.

\subsection{Rowhammer Test Bench}

Considering that FL may use different memory replacement strategies and that data diversity plays a role, this work tested different Rowhammer attack thresholds, based on two access patterns and four data patterns. Compare to single-sided access pattern, the double-sided requires less activations. However, it is also more difficult to utilize as it requires two adjacent rows for victim data. The data pattern will be the data layout between victim rows and aggressor rows. 
The tests not only reveal patterns that are more likely to induce bit flips, such as 1 hammered by 0 or 1 hammered by 0, but also demonstrate relatively more challenging patterns, such as 0 hammered by 0 or 1 hammered by 1.
Specially, we use hexadecimal strings to represent data patterns that fills the entire DRAM row. For example, \texttt{0xFF} represents 32-bit data filled with all ones, while \texttt{0x00} represents all zeros. The obtained attack thresholds are derived from the average values across all banks of the tested DRAM module.

These activation thresholds ensure that the attack induces bit-flips in at least 95\% of the DRAM rows. The inability to obtain thresholds for all rows is likely due to the presence of exceptionally resilient cells within the module, which results in extreme deviations in the average threshold. These thresholds are specifically presented in Table~\ref{tab:thresholds}. All threshold values fall within the theoretical limit of 1,309K activations and reserve enough dummy rows activations to bypass TRR. It can be observed that double-sided Rowhammer generally requires fewer activations than single-sided one. In the worst case, 265K activations are sufficient to induce bit flips.

\subsection{PPO Agent Training}
\begin{table}[!t]
\centering
\caption{Rowhammer activation thresholds under different data patterns.}
\label{tab:thresholds}
\renewcommand{\arraystretch}{1.1}
\setlength{\tabcolsep}{8pt}

\begin{tabular}{cccc}
\hline
\textbf{Victim} & \textbf{Aggressor} & \textbf{Single-sided} & \textbf{Double-sided} \\
\hline
\texttt{0xFF} & \texttt{0x00} & 185K & 115K \\
\texttt{0x00} & \texttt{0xFF} & 240K & 140K \\
\texttt{0x55} & \texttt{0x55} & 260K & 160K \\
\texttt{0xAA} & \texttt{0xAA} & 265K & 165K \\
\hline
\end{tabular}
\end{table}
We empirically evaluate the effectiveness of our proposed PPO-based adversarial framework in consistently inducing clustered and repetitive parameter updates in target FL systems under varying sparsity settings, which are suitable for Rowhammer exploitation. 

\subsubsection{\textbf{FL Setup}}
\label{subsec:exp_setup}

We selected three widely-used ASR models of varying complexity, parameter counts, and architectural patterns for evaluation and a MobileNetV3 model, see in Table~\ref{tab:models} the four models and their parameter statistics.

For all models, we simulate for two sparsity settings: 1) 0.1\% updates of moderate communication efficiency is required without severely limiting model convergence speed; and 2) 0.05\% updates that seek extreme communication efficiency conditions, simulating highly bandwidth-constrained or large-scale deployments.

The total number of simulated clients is training efficiency oriented, and scales subject to limited hardware capability and target model sizes, ranging from 20-50.

Also, to improve training efficiency, all target models are quantized version, as listed in~\ref{tab:models}.

The PPO agent was trained to generate adversarial waveforms constrained by a maximum perturbation limit ($\|\delta\|_\infty \leq 0.1$), ensuring the adversarial audio samples remain stealthy and practically imperceptible. For the ASR models, the training dataset employed was the English subset of Common Voice 17.0~\cite{commonvoice2020}, and for CV, the dataset is CIFAR10~\cite{krizhevsky2009learning}. Hyperparameters for the reward function were empirically started with $\alpha=1.0$, $\beta=0.8$, and $\gamma=0.6$, with perceptibility penalty weights set to $\lambda_1 = \lambda_2 = 0.5$ for ASR, and $\lambda = 0.8$ for CV.

\subsubsection{\textbf{Training Protocol}} 
\label{subsec:training}

The target FL system, serving as the environment for the RL agent, at each iteration, is set to run 100 communication rounds before being reset at the end. The agent generates action and observe client updates at each round. The agent is train for a total of 800 such iterations.

\begin{table}[htbp]
    \centering
    \caption{Estimated Row Activations Based on Empirical Clustering}
    \label{tab:results}
    \begin{tabular}{lcccc}
    \toprule
    \textbf{Model} & \textbf{Sparsity} & \textbf{$H_{\text{max}}$} & \textbf{RUR(\%)} & \textbf{$\mathbb{E}_{Act}$} \\
    \midrule
    Conformer-CTC-S  & 0.1\%  & 296K & 69.5 & 206K \\
    @INT4            & 0.05\% & 592K & 60.5 & 358K \\
    \midrule
    Squeezeformer-XS & 0.1\%  & 286K & 63.1 & 180K \\
    @INT4            & 0.05\% & 572K & 53.7 & 307K \\
    \midrule
    QuartzNet 5x5    & 0.1\%  & 192K & 76.0 & 145K \\
    @INT8            & 0.05\% & 384K & 60.8 & 233K \\
    \midrule
    MobileNetV3      & 0.1\%  & 444K & 63.3 & 281K \\
    @INT8            & 0.05\% & 888K & 58.3 & 518K \\
    \bottomrule
    \end{tabular}
\end{table}

\section{Results and Analysis}
\label{sec:results}

\subsection{RL Performance}

\begin{figure}[htbp]
    \centering
    \includegraphics[width=1\linewidth]{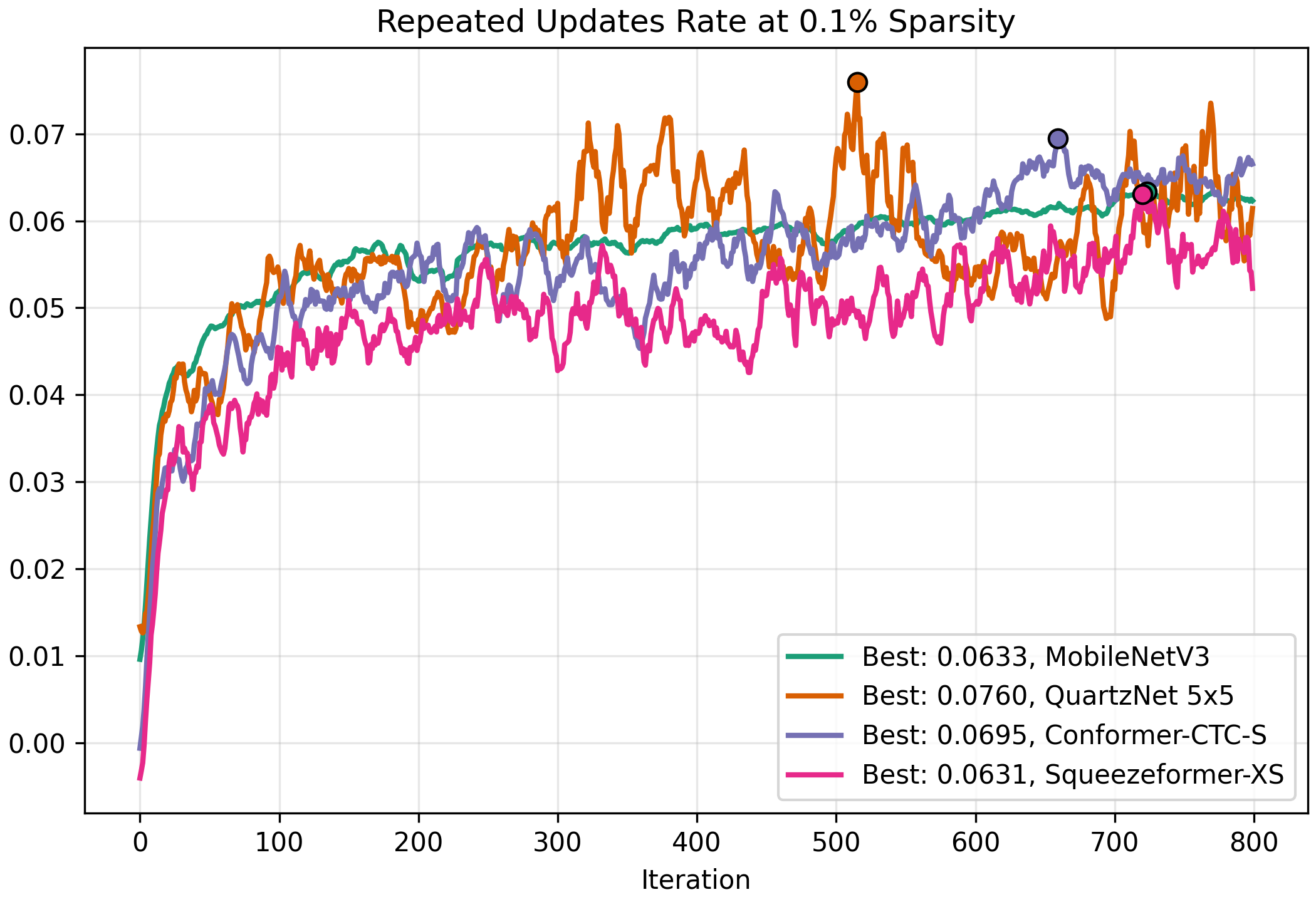}
    \includegraphics[width=1\linewidth]{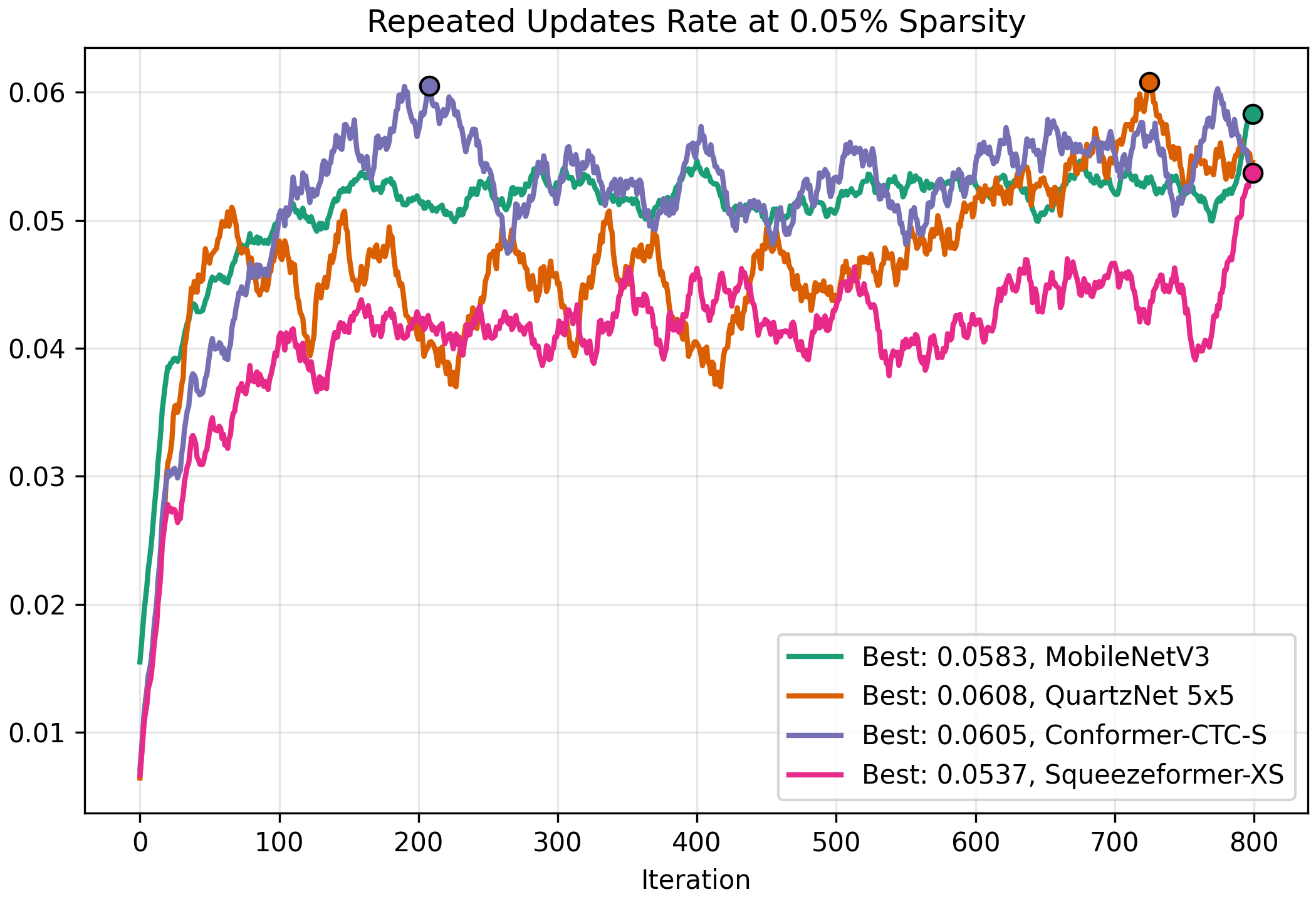}
    \caption{Repeated Update Rate of target models with sparsity of 0.1\% and 0.05\% settings, in all cases, the agent succeeded in inducing repetitive clustered updates. At sparsity of 0.1\%, attacks on all models achieved RUR above 60\% with QuartzNet 5x5 reached 76\%. At sparsity of 0.05\%, the RUR of these attacks decreased, but remain above 50\%, with the highest QuartzNet 5x5 attack at 60.8\%.}
    \label{fig:rur_0_1}
\end{figure}

Table~\ref{tab:results} and Fig.~\ref{fig:rur_0_1} present results of our PPO method across different model architectures and sparsity levels. On the target of \textbf{Repeated Update Rate (RUR)}, as summarized in Table~\ref{tab:results}, our PPO agent consistently uplifts the model update concentration at both sparsity settings. At high sparsity (0.05\%), the agent achieved RUR ranging from $53.7\%$ (Squeezeformer-XS) to $60.8\%$ (QuartzNet 5x5); while at medium sparisty (0.1\%), RUR are higher, ranging from $63.1\%$ (Squeezeformer-XS) to $76.0\%$ (QuartzNet 5x5). These results demonstrate the efficacy of our approach in inducing concentrated and stable update patterns necessary for facilitating targeted Rowhammer attacks in realistic FL settings.

\subsection{Bit-Flip Validation}

In this subsection, we jointly analyze the effectiveness of our proposed PPO-based adversarial attack framework in inducing practical Rowhammer conditions, considering both the theoretical Rowhammer thresholds obtained in Section~IV-B and the empirically measured clustering results from PPO agent training presented in Section~IV-C. Specifically, we quantify the practical feasibility of triggering Rowhammer bit-flips by correlating the observed Repeated Update Rate (RUR) with the number of sparse gradient updates performed in federated learning scenarios.

This straightforward estimation facilitates practical reasoning about the potential for successful Rowhammer attacks based on experimentally observed adversarial clustering.

\subsubsection{Estimated Row Activation Counts}
Table~\ref{tab:results} summarizes the calculated expected row activations for evaluated models at both sparsity settings. Although the QuartzNet 5x5 attack achieve slight higher RUR, the larger size and higher precision limits the $H_{\text{max}}$, and therefore, lower $\mathbb{E}_{Act}$. Among these target models, Conformer-CTC-S and MobileNetV3 have the higher $\mathbb{E}_{Act}$, both above 200K under the two sparsity settings.

\subsubsection{Estimated Bit Flips}

Refer to Table~\ref{tab:thresholds}, averaging bit-flip thresholds over different single-sided aggressor patterns, bit flips are expected when repeated activation exceeds 240K, i.e., $\overline{T_{\text{flip}}} = 240,000$.
Under the 0.1\% sparsity setting, MobileNetV3 is the only model whose estimated activation count exceeds the average threshold, indicating that Rowhammer bit flips can be reliably induced.
Conformer-CTC-S remains close to the vulnerable range and may still induce bit flips under favorable conditions, whereas Squeezeformer-XS and QuartzNet 5x5 fall below the practical threshold and are therefore unlikely to succeed.
Under the 0.05\% sparsity setting, the attack becomes more feasible overall.
Conformer-CTC-S, Squeezeformer-XS, and MobileNetV3 all exceed the average threshold, while QuartzNet 5x5 reaches 233K activations, placing it between the minimum and average threshold and indicating that bit flips can still be induced, although with greater difficulty than other three models.

\section{Discussion}

In this section, we discuss the practical implications, limitations, security considerations, and mitigation strategies related to our proposed PPO-driven Rowhammer attack against federated learning (FL) systems. 

\subsection{Practical Implications and Feasibility}

Our experiments provide empirical evidence that PPO-generated physical-domain perturbations can effectively induce clustered and repetitive parameter updates critical for successful Rowhammer exploitation in federated learning scenarios. 

However, practical real-world deployment may introduce complexities such as partial system knowledge, environmental variability, or limited precision in physical perturbation generation. Despite these uncertainties, our joint evaluation clearly demonstrates that the activation counts achieved by our PPO-based attacks are realistically within the threshold ranges required to trigger Rowhammer attacks, particularly under favorable DRAM configurations or minor system optimizations (e.g., higher memory bandwidth or extended refresh intervals). Therefore, our results underscore a significant and realistic security risk to federated learning systems utilizing common performance optimization practices.

\subsection{Limitations and Open Challenges}

We acknowledge two primary limitations on the practical applicability of the proposed remote Rowhammer attack:

\begin{enumerate}
    \item \textbf{Opacity of Target FL Setups:}
    Real-world federated systems typically obscure internal details such as aggregation policies, memory management, and system optimizations from external attackers. Without explicit feedback or partial system knowledge, the current reinforcement learning framework might experience significantly reduced efficiency and effectiveness. Future research could address this limitation by developing RL agents with highly sparse rewards or inferable rewards, or transferable adversarial perturbation techniques.

    \item \textbf{Lack of Control over Specific Victim Rows:}
    The proposed indirect Rowhammer attack inherently lacks precise control over the specific DRAM rows affected by induced bit-flips, making outcomes unpredictable and potentially transient. This uncertainty reduces the reliability of achieving specific objectives (e.g., targeted data corruption or privilege escalation). Future research should explore advanced side-channel techniques or hardware-software co-design approaches to improve the precision and predictability of the Rowhammer effects induced by physical-domain perturbations.
\end{enumerate}

\subsection{Mitigation Measures}

Based on our findings, we propose several practical defensive measures to mitigate the identified threat:

\begin{itemize}
    \item \textbf{Adversarial Input Defense:} From the sofware prospective, as a variant of adversarial attack, defense measures, such as input space transformation, feature denoising and adversarial input detection and rejection, are also effective. The actual effectiveness of these measures depends on the tolerance of delay on clients end, as well as adapting settings to the variant of adversarial attack. Input space transformation and feature denoising can filter adversarial perturbations of lower magnitudes, while perturbations of higher magnitudes are relatively easier to detect and reject.

    \item \textbf{Hardware-Level Countermeasures:}
    From the hardware perspective, upgrading to DRAM modules with enhanced Rowhammer resilience, such as those with advanced error correction, higher refresh frequency, offers direct defense against Rowhammer exploitation attempts of all sources.

    \item \textbf{Systematic Security Analysis on Deployment:} on deployment of FL systems,  all optimization techniques, software or hardware based ones, should be systematically assessed across both software execution and hardware behavior, to prevent unintentional susceptibility to threat vectors such as our proposed vector when improving computational and communication efficiency. To reduce the feasibility of the proposed attack vector, practitioners should evaluate optimization trade-offs not only from a performance perspective but also with respect to their security implications, particularly in sensitive or critical infrastructure environments.
\end{itemize}

\subsection{Broader Impact and Future Directions}

Our study demonstrates a need for interdisciplinary approaches combining hardware security, adversarial machine learning, and federated learning systems design. By integrating hardware-level vulnerabilities into FL threat modelling, we advocate a holistic view that proactively identifies, assesses, and mitigates emerging threats, leading to the following future research directions:
\begin{itemize}
    \item Exploring hardware-software joint-design approaches that mitigate Rowhammer risks without severely compromising computational efficiency.
    \item Extending our approach to diverse model architectures, and real-world applications through stronger agents that can process side channel information and learn with highly sparse rewards in real-world applications to enhance the attack.
\end{itemize}

By addressing these open challenges and continuing to bridge security gaps between hardware vulnerabilities and distributed learning systems, we aim to contribute toward more secure, dependable, and resilient federated learning deployments.

\section{Conclusion}

In this paper, we introduced and comprehensively evaluated a novel Rowhammer attack vector targeting federated learning (FL) systems, exploiting client-side physical interference orchestrated by a Proximal Policy Optimization (PPO)-based reinforcement learning agent. Unlike traditional Rowhammer attacks, our method leverages adversarial acoustic or electromagnetic perturbations at client sensors, indirectly inducing clustered and repetitive parameter updates on FL servers. Our empirical results demonstrate the feasibility of our propose attack vector, effectively increasing the frequency of repeated DRAM row activations for triggering Rowhammer bit-flips.

Through extensive evaluation and analysis across diverse Automatic Speech Recognition (ASR) architectures as well as a CIFAR10 classification MobileNetV3, we demonstrated that the proposed PPO approach consistently achieves repeated clustered updates. The evaluation highlighted practical feasibility, with achieved activation counts realistically approaching Rowhammer thresholds under favorable conditions, emphasizing the tangible security threat posed by this attack vector.

Furthermore, we analyzed the security implications associated with common FL optimizations such as sparse updates, pinned memory, and RDMA, illustrating how performance-oriented design choices can unintentionally expose systems to hardware-level vulnerabilities. We outlined practical mitigation strategies, including randomized memory allocation, adversarial input detection, and controlled usage of optimization techniques, as proactive defensive measures to counteract this emerging threat.

Our findings underscore the necessity for interdisciplinary threat modeling, encompassing both software and hardware domains, to ensure the secure and dependable deployment of federated learning systems. 

\bibliographystyle{IEEEtran}
\bibliography{IEEEabrv, reference}

\begin{thebibliography}{10}
\providecommand{\url}[1]{#1}
\csname url@samestyle\endcsname
\providecommand{\newblock}{\relax}
\providecommand{\bibinfo}[2]{#2}
\providecommand{\BIBentrySTDinterwordspacing}{\spaceskip=0pt\relax}
\providecommand{\BIBentryALTinterwordstretchfactor}{4}
\providecommand{\BIBentryALTinterwordspacing}{\spaceskip=\fontdimen2\font plus
\BIBentryALTinterwordstretchfactor\fontdimen3\font minus \fontdimen4\font\relax}
\providecommand{\BIBforeignlanguage}[2]{{%
\expandafter\ifx\csname l@#1\endcsname\relax
\typeout{** WARNING: IEEEtran.bst: No hyphenation pattern has been}%
\typeout{** loaded for the language `#1'. Using the pattern for}%
\typeout{** the default language instead.}%
\else
\language=\csname l@#1\endcsname
\fi
#2}}
\providecommand{\BIBdecl}{\relax}
\BIBdecl

\bibitem{mcmahan2017communication}
B.~McMahan, E.~Moore, D.~Ramage, S.~Hampson, and B.~A. y~Arcas, ``Communication-efficient learning of deep networks from decentralized data,'' in \emph{Artificial intelligence and statistics}.\hskip 1em plus 0.5em minus 0.4em\relax PMLR, 2017, pp. 1273--1282.

\bibitem{kairouz2021advances}
P.~Kairouz, H.~B. McMahan, B.~Avent, A.~Bellet, M.~Bennis, A.~N. Bhagoji, K.~Bonawitz, Z.~Charles, G.~Cormode, R.~Cummings \emph{et~al.}, ``Advances and open problems in federated learning,'' \emph{Foundations and trends{\textregistered} in machine learning}, vol.~14, no. 1--2, 2021.

\bibitem{kim2014flipping}
Y.~Kim, R.~Daly, J.~Kim, C.~Fallin, J.~H. Lee, D.~Lee, C.~Wilkerson, K.~Lai, and O.~Mutlu, ``Flipping bits in memory without accessing them: An experimental study of dram disturbance errors,'' \emph{ACM SIGARCH Computer Architecture News}, vol.~42, no.~3, pp. 361--372, 2014.

\bibitem{frigo2020trrespass}
P.~Frigo, E.~Vannacc, H.~Hassan, V.~Van Der~Veen, O.~Mutlu, C.~Giuffrida, H.~Bos, and K.~Razavi, ``Trrespass: Exploiting the many sides of target row refresh,'' in \emph{2020 IEEE Symposium on Security and Privacy (SP)}.\hskip 1em plus 0.5em minus 0.4em\relax IEEE, 2020, pp. 747--762.

\bibitem{jattke2024zenhammer}
P.~Jattke, M.~Wipfli, F.~Solt, M.~Marazzi, M.~B{\"o}lcskei, and K.~Razavi, ``Zenhammer: Rowhammer attacks on amd zen-based platforms,'' in \emph{33rd USENIX Security Symposium (USENIX Security 2024)}, 2024.

\bibitem{seaborn2015exploiting}
M.~Seaborn and T.~Dullien, ``Exploiting the dram rowhammer bug to gain kernel privileges,'' \emph{Black Hat}, vol.~15, no.~71, p.~2, 2015.

\bibitem{gruss2016rowhammer}
D.~Gruss, C.~Maurice, and S.~Mangard, ``Rowhammer. js: A remote software-induced fault attack in javascript,'' in \emph{Detection of Intrusions and Malware, and Vulnerability Assessment: 13th International Conference, DIMVA 2016, San Sebasti{\'a}n, Spain, July 7-8, 2016, Proceedings 13}.\hskip 1em plus 0.5em minus 0.4em\relax Springer, 2016, pp. 300--321.

\bibitem{tatar2018throwhammer}
A.~Tatar, R.~K. Konoth, E.~Athanasopoulos, C.~Giuffrida, H.~Bos, and K.~Razavi, ``Throwhammer: Rowhammer attacks over the network and defenses,'' in \emph{2018 USENIX Annual Technical Conference (USENIX ATC 18)}, 2018, pp. 213--226.

\bibitem{eykholt2018robust}
K.~Eykholt, I.~Evtimov, E.~Fernandes, B.~Li, A.~Rahmati, C.~Xiao, A.~Prakash, T.~Kohno, and D.~Song, ``Robust physical-world attacks on deep learning visual classification,'' in \emph{Proceedings of the IEEE conference on computer vision and pattern recognition}, 2018.

\bibitem{carlini2018audio}
N.~Carlini and D.~Wagner, ``Audio adversarial examples: Targeted attacks on speech-to-text,'' in \emph{2018 IEEE security and privacy workshops (SPW)}.\hskip 1em plus 0.5em minus 0.4em\relax IEEE, 2018, pp. 1--7.

\bibitem{mutlu2019rowhammer}
O.~Mutlu and J.~S. Kim, ``Rowhammer: A retrospective,'' \emph{IEEE Transactions on Computer-Aided Design of Integrated Circuits and Systems}, vol.~39, no.~8, pp. 1555--1571, 2019.

\bibitem{hassan2021uncovering}
H.~Hassan, Y.~C. Tugrul, J.~S. Kim, V.~Van~der Veen, K.~Razavi, and O.~Mutlu, ``Uncovering in-dram rowhammer protection mechanisms: A new methodology, custom rowhammer patterns, and implications,'' in \emph{MICRO-54: 54th Annual IEEE/ACM International Symposium on Microarchitecture}, 2021, pp. 1198--1213.

\bibitem{de2021smash}
F.~de~Ridder, P.~Frigo, E.~Vannacci, H.~Bos, C.~Giuffrida, and K.~Razavi, ``$\{$SMASH$\}$: Synchronized many-sided rowhammer attacks from $\{$JavaScript$\}$,'' in \emph{30th USENIX Security Symposium (USENIX Security 21)}, 2021, pp. 1001--1018.

\bibitem{jiang2021trrscope}
Y.~Jiang, H.~Zhu, H.~Shan, X.~Guo, X.~Zhang, and Y.~Jin, ``Trrscope: Understanding target row refresh mechanism for modern ddr protection,'' in \emph{2021 IEEE International Symposium on Hardware Oriented Security and Trust (HOST)}.\hskip 1em plus 0.5em minus 0.4em\relax IEEE, 2021, pp. 239--247.

\bibitem{aweke2016anvil}
Z.~B. Aweke, S.~F. Yitbarek, R.~Qiao, R.~Das, M.~Hicks, Y.~Oren, and T.~Austin, ``Anvil: Software-based protection against next-generation rowhammer attacks,'' \emph{ACM SIGPLAN Notices}, vol.~51, no.~4, 2016.

\bibitem{cojocar2019exploiting}
L.~Cojocar, K.~Razavi, C.~Giuffrida, and H.~Bos, ``Exploiting correcting codes: On the effectiveness of ecc memory against rowhammer attacks,'' in \emph{2019 IEEE Symposium on Security and Privacy (SP)}.\hskip 1em plus 0.5em minus 0.4em\relax IEEE, 2019.

\bibitem{cojocar2020we}
L.~Cojocar, J.~Kim, M.~Patel, L.~Tsai, S.~Saroiu, A.~Wolman, and O.~Mutlu, ``Are we susceptible to rowhammer? an end-to-end methodology for cloud providers,'' in \emph{2020 IEEE symposium on security and privacy (SP)}.\hskip 1em plus 0.5em minus 0.4em\relax IEEE, 2020, pp. 712--728.

\bibitem{xiao2016one}
Y.~Xiao, X.~Zhang, Y.~Zhang, and R.~Teodorescu, ``One bit flips, one cloud flops:$\{$Cross-VM$\}$ row hammer attacks and privilege escalation,'' in \emph{25th USENIX security symposium}, 2016, pp. 19--35.

\bibitem{kang2024sledgehammer}
I.~Kang, W.~Wang, J.~Kim, S.~van Schaik, Y.~Tobah, D.~Genkin, A.~Kwong, and Y.~Yarom, ``$\{$SledgeHammer$\}$: Amplifying rowhammer via bank-level parallelism,'' in \emph{33rd USENIX Security Symposium (USENIX Security 24)}, 2024, pp. 1597--1614.

\bibitem{pessl2016drama}
P.~Pessl, D.~Gruss, C.~Maurice, M.~Schwarz, and S.~Mangard, ``$\{$DRAMA$\}$: Exploiting $\{$DRAM$\}$ addressing for $\{$Cross-CPU$\}$ attacks,'' in \emph{25th USENIX security symposium (USENIX security 16)}, 2016, pp. 565--581.

\bibitem{qiao2016new}
R.~Qiao and M.~Seaborn, ``A new approach for rowhammer attacks,'' in \emph{2016 IEEE international symposium on hardware oriented security and trust (HOST)}.\hskip 1em plus 0.5em minus 0.4em\relax IEEE, 2016, pp. 161--166.

\bibitem{jang2017sgx}
Y.~Jang, J.~Lee, S.~Lee, and T.~Kim, ``Sgx-bomb: Locking down the processor via rowhammer attack,'' in \emph{Proceedings of the 2nd Workshop on System Software for Trusted Execution}, 2017, pp. 1--6.

\bibitem{lipp2020nethammer}
M.~Lipp, M.~Schwarz, L.~Raab, L.~Lamster, M.~T. Aga, C.~Maurice, and D.~Gruss, ``Nethammer: Inducing rowhammer faults through network requests,'' in \emph{2020 IEEE European Symposium on Security and Privacy Workshops (EuroS\&PW)}.\hskip 1em plus 0.5em minus 0.4em\relax IEEE, 2020, pp. 710--719.

\bibitem{cuda_guide}
{NVIDIA Corporation}, ``{CUDA C++ Programming Guide},'' \url{https://docs.nvidia.com/cuda/cuda-c-programming-guide/index.html}, 2024, accessed: 2026-04-15.

\bibitem{linux_thp}
{Linux Kernel}, ``{Transparent Huge Pages},'' \url{https://www.kernel.org/doc/html/latest/admin-guide/mm/transhuge.html}, 2024, accessed: 2026-04-15.

\bibitem{zhang2024fedrdma}
Z.~Zhang, D.~Cai, Y.~Zhang, M.~Xu, S.~Wang, and A.~Zhou, ``Fedrdma: Communication-efficient cross-silo federated llm via chunked rdma transmission,'' in \emph{Proceedings of the 4th Workshop on Machine Learning and Systems}, 2024, pp. 126--133.

\bibitem{szegedy2013intriguing}
C.~Szegedy, ``Intriguing properties of neural networks,'' \emph{arXiv1312.6199}, 2013.

\bibitem{goodfellow2014explaining}
I.~J. Goodfellow, J.~Shlens, and C.~Szegedy, ``Explaining and harnessing adversarial examples,'' \emph{arXiv1412.6572}, 2014.

\bibitem{kurakin2018adversarial}
A.~Kurakin, I.~J. Goodfellow, and S.~Bengio, ``Adversarial examples in the physical world,'' in \emph{Artificial intelligence safety and security}.\hskip 1em plus 0.5em minus 0.4em\relax Chapman and Hall/CRC, 2018, pp. 99--112.

\bibitem{brown2017adversarial}
T.~B. Brown, D.~Man{\'e}, A.~Roy, M.~Abadi, and J.~Gilmer, ``Adversarial patch,'' \emph{arXiv1712.09665}, 2017.

\bibitem{evtimov2017robust}
I.~Evtimov, K.~Eykholt, E.~Fernandes, T.~Kohno, B.~Li, A.~Prakash, A.~Rahmati, and D.~Song, ``Robust physical-world attacks on machine learning models,'' \emph{arXiv1707.08945}, vol.~2, no.~3, p.~4, 2017.

\bibitem{athalye2018synthesizing}
A.~Athalye, L.~Engstrom, A.~Ilyas, and K.~Kwok, ``Synthesizing robust adversarial examples,'' in \emph{International conference on machine learning}.\hskip 1em plus 0.5em minus 0.4em\relax PMLR, 2018, pp. 284--293.

\bibitem{duan2020adversarial}
R.~Duan, X.~Ma, Y.~Wang, J.~Bailey, A.~K. Qin, and Y.~Yang, ``Adversarial camouflage: Hiding physical-world attacks with natural styles,'' in \emph{Proceedings of the IEEE/CVF conference on computer vision and pattern recognition}, 2020, pp. 1000--1008.

\bibitem{yuan2018commandersong}
X.~Yuan, Y.~Chen, Y.~Zhao, Y.~Long, X.~Liu, K.~Chen, S.~Zhang, H.~Huang, X.~Wang, and C.~A. Gunter, ``$\{$CommanderSong$\}$: a systematic approach for practical adversarial voice recognition,'' in \emph{27th USENIX security symposium (USENIX security 18)}, 2018.

\bibitem{zhou2019hidden}
M.~Zhou, Z.~Qin, X.~Lin, S.~Hu, Q.~Wang, and K.~Ren, ``Hidden voice commands: Attacks and defenses on the vcs of autonomous driving cars,'' \emph{IEEE Wireless Communications}, vol.~26, no.~5, 2019.

\bibitem{gulati2020conformer}
A.~Gulati, J.~Qin, C.-C. Chiu, N.~Parmar, Y.~Zhang, J.~Yu, W.~Han, S.~Wang, Z.~Zhang, Y.~Wu \emph{et~al.}, ``Conformer: Convolution-augmented transformer for speech recognition,'' \emph{arXiv2005.08100}, 2020.

\bibitem{kim2022squeezeformer}
S.~Kim, A.~Gholami, A.~Shaw, N.~Lee, K.~Mangalam, J.~Malik, M.~W. Mahoney, and K.~Keutzer, ``Squeezeformer: An efficient transformer for automatic speech recognition,'' \emph{Advances in Neural Information Processing Systems}, vol.~35, pp. 9361--9373, 2022.

\bibitem{kriman2020quartznet}
S.~Kriman, S.~Beliaev, B.~Ginsburg, J.~Huang, O.~Kuchaiev, V.~Lavrukhin, R.~Leary, J.~Li, and Y.~Zhang, ``Quartznet: Deep automatic speech recognition with 1d time-channel separable convolutions,'' in \emph{ICASSP 2020-2020 IEEE International Conference on Acoustics, Speech and Signal Processing (ICASSP)}.\hskip 1em plus 0.5em minus 0.4em\relax IEEE, 2020, pp. 6124--6128.

\bibitem{howard2019searching}
A.~Howard, M.~Sandler, G.~Chu, L.-C. Chen, B.~Chen, M.~Tan, W.~Wang, Y.~Zhu, R.~Pang, V.~Vasudevan \emph{et~al.}, ``Searching for mobilenetv3,'' in \emph{Proceedings of the IEEE/CVF international conference on computer vision}, 2019, pp. 1314--1324.

\bibitem{commonvoice2020}
R.~Ardila, M.~Branson, K.~Davis, M.~Henretty, M.~Kohler, J.~Meyer, R.~Morais, L.~Saunders \emph{et~al.}, ``Common voice: A massively-multilingual speech corpus,'' in \emph{Proceedings of the 12th Conference on Language Resources and Evaluation (LREC 2020)}, 2020.

\bibitem{krizhevsky2009learning}
A.~Krizhevsky, G.~Hinton \emph{et~al.}, ``Learning multiple layers of features from tiny images,'' University of Toronto, Tech. Rep., 2009.

\bibitem{bengio2009curriculum}
\BIBentryALTinterwordspacing
Y.~Bengio, J.~Louradour, R.~Collobert, and J.~Weston, ``Curriculum learning,'' in \emph{Proceedings of the 26th Annual International Conference on Machine Learning}, ser. ICML '09.\hskip 1em plus 0.5em minus 0.4em\relax New York, NY, USA: Association for Computing Machinery, 2009, p. 41–48. [Online]. Available: \url{https://doi.org/10.1145/1553374.1553380}
\BIBentrySTDinterwordspacing

\bibitem{he2016deep}
K.~He, X.~Zhang, S.~Ren, and J.~Sun, ``Deep residual learning for image recognition,'' in \emph{Proceedings of the IEEE conference on computer vision and pattern recognition}, 2016, pp. 770--778.

\bibitem{olgun2023dram}
A.~Olgun, H.~Hassan, A.~G. Ya{\u{g}}l{\i}k{\c{c}}{\i}, Y.~C. Tu{\u{g}}rul, L.~Orosa, H.~Luo, M.~Patel, O.~Ergin, and O.~Mutlu, ``Dram bender: An extensible and versatile fpga-based infrastructure to easily test state-of-the-art dram chips,'' \emph{IEEE Transactions on Computer-Aided Design of Integrated Circuits and Systems}, vol.~42, no.~12, pp. 5098--5112, 2023.

\bibitem{alveo}
\BIBentryALTinterwordspacing
``Alveo u200 and u250 data center accelerator cards data sheet (ds962),'' accessed: 28-Feb-2025. [Online]. Available: \url{https://docs.amd.com/r/en-US/ds962-u200-u250}
\BIBentrySTDinterwordspacing

\bibitem{memory}
\BIBentryALTinterwordspacing
``Micron ddr4 sdram rdimm mta18asf2g72pz-2g3b1 datasheet,'' accessed: 28-Feb-2025. [Online]. Available: \url{https://www.mouser.co.uk/datasheet/2/671/asf18c2gx72pz-3079314.pdf}
\BIBentrySTDinterwordspacing

\end{thebibliography}

\begin{IEEEbiography}[{\includegraphics[width=1in,height=1.25in,clip,keepaspectratio]{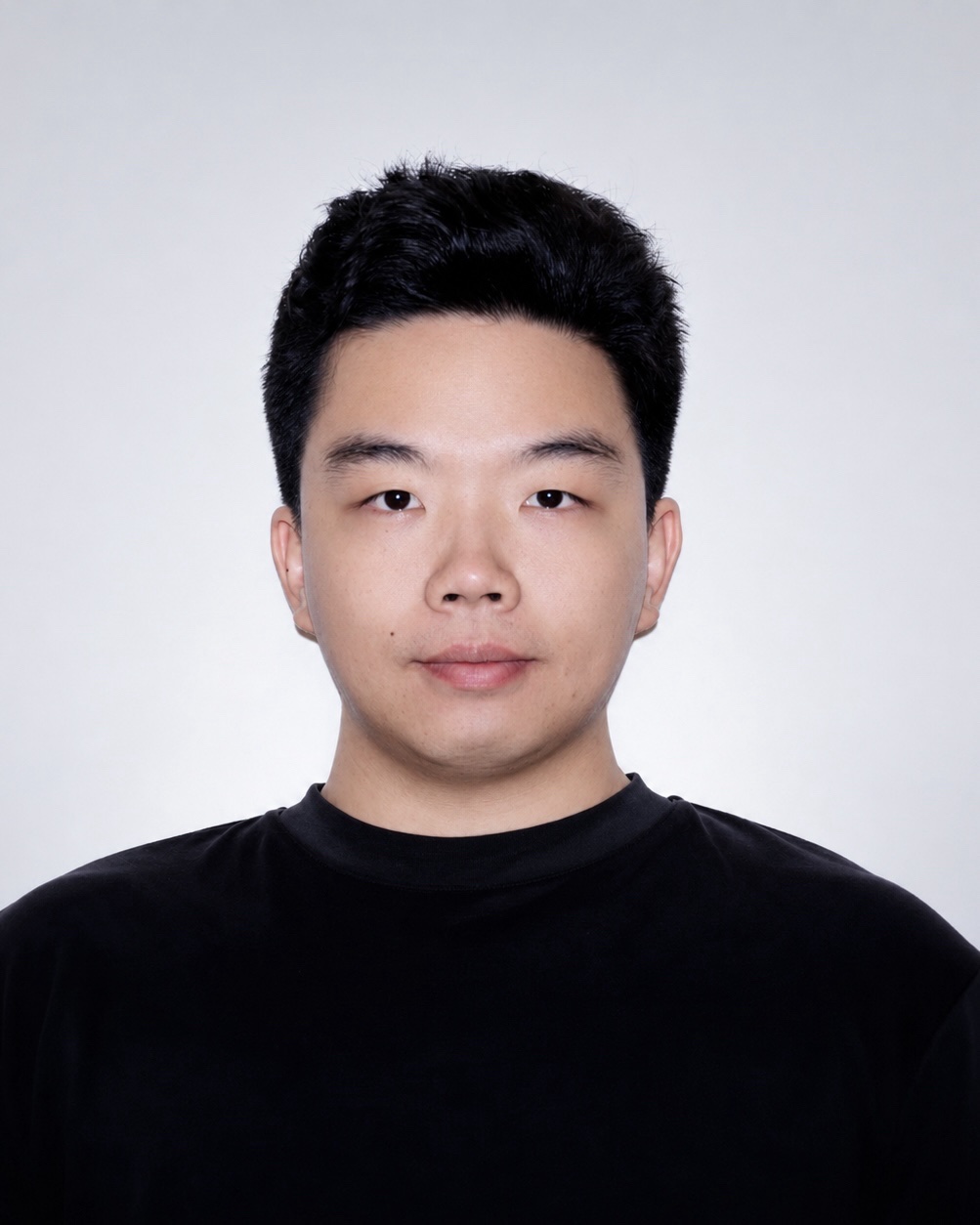}}]
{Jinsheng Yuan}
received his PhD degree from the Cranfield University in 2025. He is currently a research fellow at Centre for Assured and Connected Autonomy, Faculty of Engineering and Applied Sciences, Cranfield University, UK. His research interests include distributed learning, cross-layer security and adversarial AI.
\end{IEEEbiography}

\begin{IEEEbiography}[{\includegraphics[width=1in,height=1.25in,clip,keepaspectratio]{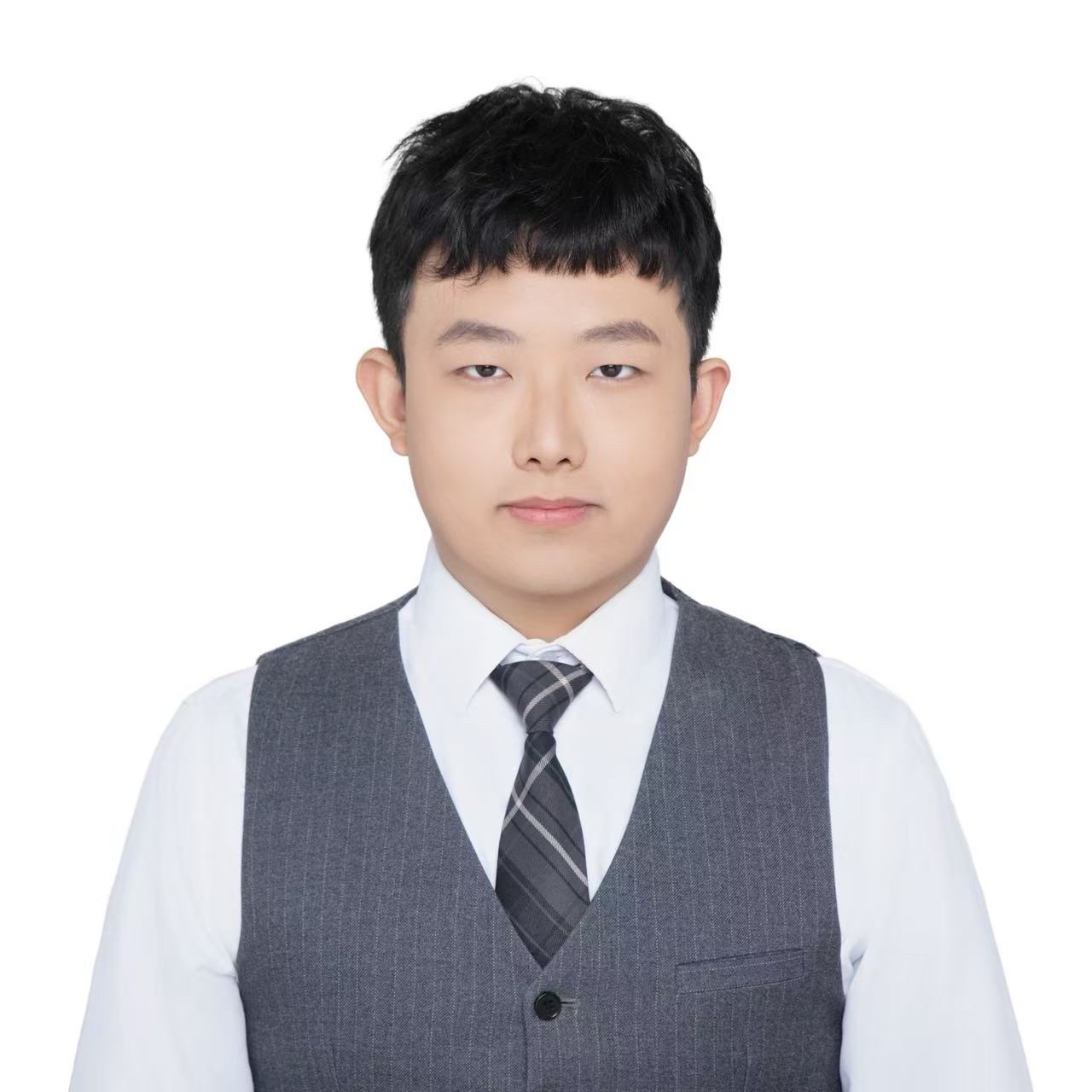}}]
{Yuhang Hao}
received the M.Sc. degree in Cyber Security from Queen's University Belfast in 2023. He is currently a PhD student with the Centre for Secure Information Technologies,
Queen’s University of Belfast.
His research interests include hardware security, approximate computing, and memory security. His current research focuses on Rowhammer attacks and their interaction with approximate computing systems.
\end{IEEEbiography}

\begin{IEEEbiography}[{\includegraphics[width=1in,height=1.25in,clip,keepaspectratio]{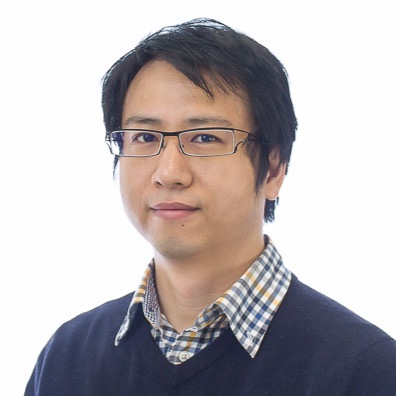}}]
{Yun Wu}
(Member, IEEE) 
is currently a Lecturer (Assistant Professor) in computing systems with the School of Electronics, Electrical Engineering and Computer Science, Queen's University Belfast, Belfast, U.K.

He received the Ph.D. degree in electronic, electrical and information engineering from Queen's University Belfast in 2013. He subsequently held research positions with Queen's University Belfast, Cranfield University, and Heriot-Watt University, U.K., before returning to Queen's in 2023. His research interests include approximate computing, reconfigurable accelerator architectures, energy-efficient embedded systems, and efficient and trustworthy AI for autonomous systems.

Dr. Wu is a Member of IEEE and of the EPSRC Peer Review College. He is a member of the VLSI Systems and Applications Technical Committee (VSA-TC) of the IEEE Circuits and Systems Society.
\end{IEEEbiography}

\begin{IEEEbiography}[{\includegraphics[width=1in,height=1.25in,clip,keepaspectratio]{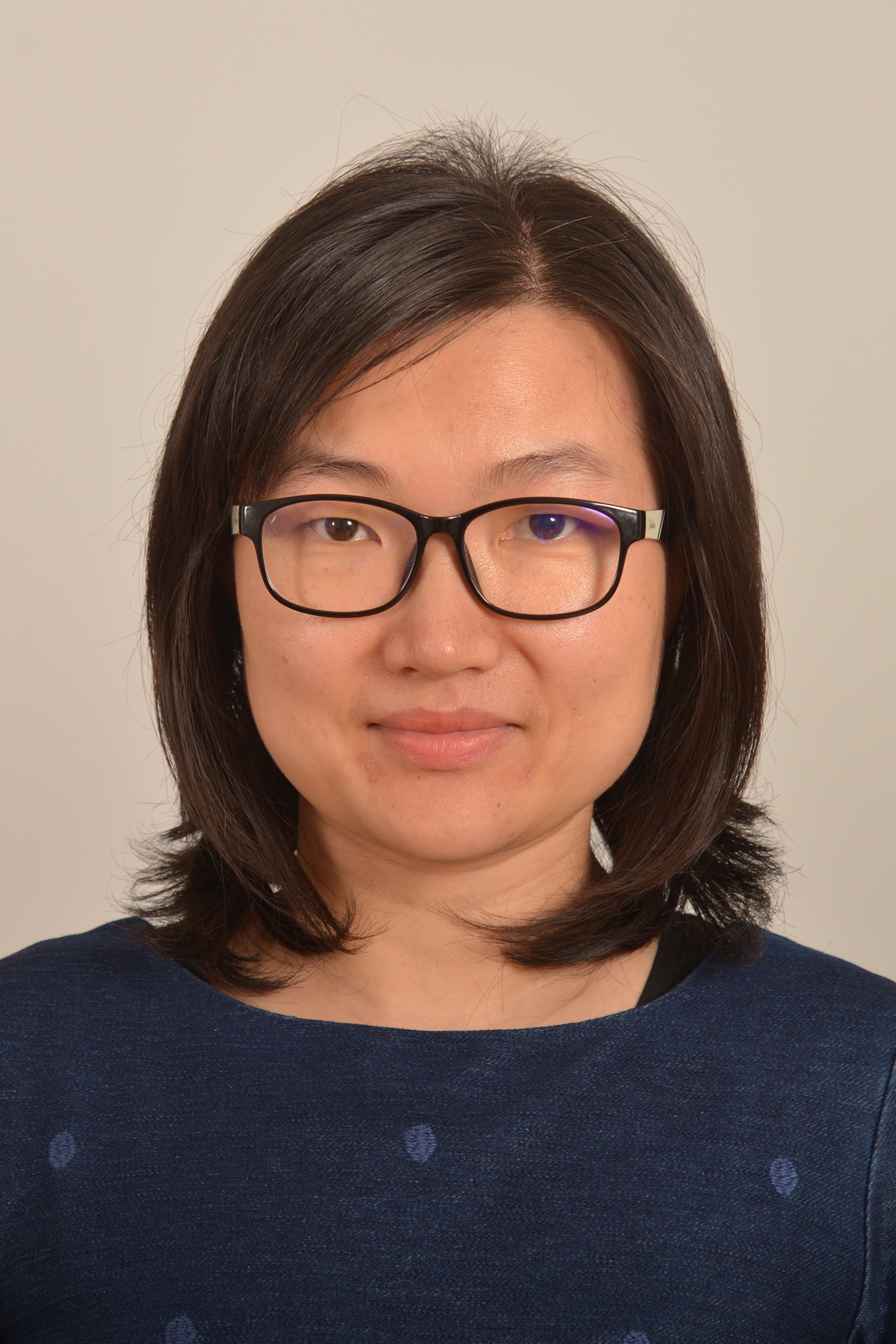}}]
{Chongyan Gu}
(Senior Member, IEEE)
received the Ph.D. degree from Queen's University Belfast, Belfast, U.K., in 2016. She is currently a Senior Lecturer (Associate Professor) in the School of EEECS at Queen's University Belfast. 

She has been awarded the EPSRC New Investigator Award in 2022. Her research into physical unclonable function (PUF) has been utilized as part of a security architecture for electronic vehicle (EV) charging systems, licensed by LG-CNS, South Korea. Her team was the overall winner of INVENT 2015, a competition to accelerate the commercialisation of innovative ideas. She serves as an Associate Editor of IEEE Transactions on Emerging Technologies in Computing (TETC) and Guest Editor of IEEE Transactions on Circuits and Systems – I (TCAS-I). She also serves on Technical Programme Committees of the DAC, ISCAS, ASP-DAC, SOCC and AsianHOST conferences. She served on the Organising Committee of AsianHOST and the workshop and tutorial chair of the 32nd FPL 2022 conference. Her co-authored pioneering research paper (invited) in this research topic has been published by Proceedings of the IEEE (impact factor: 10.694). She was invited to give keynote/tutorial talks to international conferences, such as, HiPEAC 2023 and IEEE ASP-DAC 2020. Her current research interests include PUFs, security in/for approximate computing, true random number generator (TRNGs), hardware Trojan detection and machine learning attacks. 

She is an IEEE Senior member since 2021.
\end{IEEEbiography}

\begin{IEEEbiography}[{\includegraphics[width=1in,height=1.25in,clip,keepaspectratio]{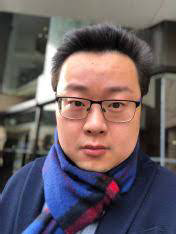}}]
{Weisi Guo}
(Senior Member, IEEE)
received his PhD degree from the University of Cambridge, UK in 2011. He was an Associate Professor at the University of Warwick (2012-19) and Turing Fellow (2017-19). He is currently a full Professor and head of the Human Machine Intelligence Group at Cranfield University, UK. He was a winner of the IET Innovation Award. His research interests focus on networks, data science, and autonomy.
\end{IEEEbiography}

\end{document}